\documentclass[preprint,11pt,authoryear]{elsarticle}

\usepackage{amssymb}
\usepackage{tikz}
\usepackage{amsmath}
\usepackage[a4paper,margin=1.5cm]{geometry}
\usepackage{graphicx}
\usepackage{natbib}
\usepackage{subcaption}
\usepackage{lineno}
\usepackage{multirow}  
\usepackage{array}
\usepackage{xcolor}      
\usepackage{lmodern}      
\usepackage[T1]{fontenc}  
\newcommand{\cyr}[1]{{\color{black}#1}}

\usepackage[unicode=false]{hyperref}
\begin{document}
\journal{Solar Energy}
\begin{frontmatter}

\title{On the Importance of Clearsky Model in Short-Term Solar Radiation Forecasting}

\author[label1]{Cyril Voyant}
\author[label2,label3]{Milan Despotovic}
\author[label3]{Gilles Notton}
\author[label1]{Yves-Marie Saint-Drenan}
\author[label3]{Mohammed Asloune}
\author[label3]{Luis Garcia-Gutierrez}
\affiliation [label1] {organization={OIE Laboratory},
            addressline={Mines-PSL}, 
            city={Sophia-Antipolis},
            postcode={F-06904}, 
            state={Antibes},
            country={France}}
\affiliation [label2] {organization={Faculty of Engineering},
            addressline={University of Kragujevac}, 
            city={Kragujevac},
            postcode={34000}, 
            country={Serbia}}            
\affiliation [label3] {organization={SPE Laboratory, UMR CNRS 6134},
            addressline={University of Corsica Pasquale Paoli}, 
            city={Ajaccio},
            postcode={20000 }, 
            state={Corsica},
            country={France}}            

\begin{abstract}
Clearsky models are widely used in solar energy for many applications such as quality control, resource assessment, satellite-base irradiance estimation and forecasting. However, their use in forecasting and nowcasting is associated with a number of challenges. Synchronization errors, reliance on the Clearsky index (ratio of the global horizontal irradiance to its cloud-free counterpart) and high sensitivity of the clearsky model to errors in aerosol optical depth at low solar elevation limit their added value in real-time applications. This paper explores the feasibility of short-term forecasting without relying on a clearsky model. We propose a Clearsky-Free forecasting approach using Extreme Learning Machine (\cyr{\texttt{ELM}}) models. \cyr{\texttt{ELM}} learns daily periodicity and local variability directly from raw Global Horizontal Irradiance (\cyr{\texttt{\cyr{\texttt{GHI}}}}) data. It eliminates the need for Clearsky normalization, simplifying the forecasting process and improving scalability. Our approach is a non-linear adaptative statistical method that implicitely learns the irradiance in cloud-free conditions removing the need for an clear-sky model and the related operational issues. Deterministic and probabilistic results are compared to traditional benchmarks, including ARMA with \cyr{\texttt{McClear}}-generated Clearsky data and quantile regression for probabilistic forecasts. \cyr{\texttt{ELM}} matches or outperforms these methods, providing accurate predictions and robust uncertainty quantification. This approach offers a simple, efficient solution for real-time solar forecasting. By overcoming the stationarization process limitations based on usual multiplicative scheme Clearsky models, it provides a flexible and reliable framework for modern energy systems.
\end{abstract}

\begin{graphicalabstract}
\includegraphics[width=\linewidth]{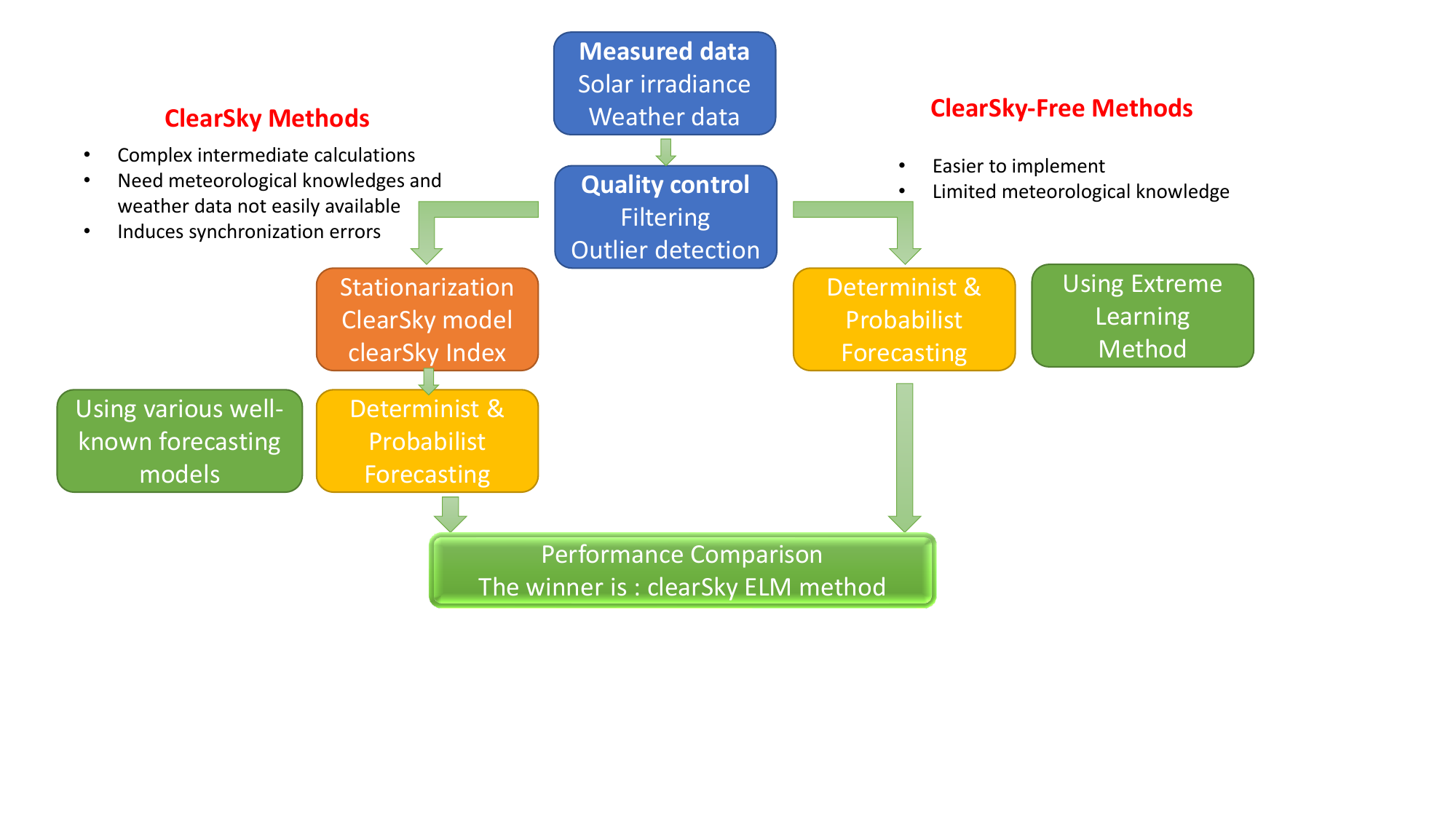}
\end{graphicalabstract}

\begin{highlights}
\item A novel method using raw \cyr{\texttt{GHI}} and machine learning, outperforming Clearsky-Based models.
\item Clearsky-Free and Clearsky-Based models in comparison concerning deterministic and probabilistic concerns.
\item Avoiding the Clearsky Index reduces dependency on complex intermediate calculations.
\item Enables smarter grids, energy trading, and renewable integration.
\end{highlights}

\begin{keyword}
 Solar Irradiance Forecasting \sep Clearsky Models \sep Clearsky-Free Approach \sep Extreme Learning Machine (\cyr{\texttt{ELM}})
\end{keyword}

\end{frontmatter}

\section{Introduction}
Solar irradiance forecasting is essential for integrating solar energy into grids, ensuring stability, facilitating the energy balance between production and consumption and addressing variability \citep{ren2015review}. Precise forecasts optimize production, storage, and distribution, supporting the transition to sustainable energy systems \citep{yang2018review, lauret2012bayesianmodelcommitteeapproach}. Despite significant advancements in the field, forecasting accuracy remains constrained by the complex and dynamic nature of solar irradiance. The interaction between atmospheric conditions and temporal variability introduces uncertainties that complicate real-time applications and long-term planning. This study explores these uncertainties and investigates innovative solutions to enhance scalability, minimize reliance on traditional normalization methods, and improve overall forecasting reliability.

\subsection{Importance and Challenges of Solar Irradiance Forecasting}
Solar energy is expected to reach over 1,500 $GW_{p}$ of PV capacity by 2030 tripling in just 10 years, which highlights the critical need for accurate forecasting to balance energy supply, optimize grid performance, and decrease dependence on fossil fuels \citep{iea2021solar}. Real-time grid management and energy markets require accurate short-term predictions (range from about 30 min to 5-6 hours) since errors in these forecasts can result in inefficiencies and increased costs \citep{diagne2013review, voyant2017machine}.
However, forecasting global solar irradiance presents significant challenges due to the inherent variability and non-stationarity of solar irradiance data. In order to solve these challenges, irradiance data are often normalized. One way to normalize solar irradiance data involves utilizing extraterrestrial irradiance that is defined as solar irradiance at the topmost layer of the Earth's atmosphere. This extraterrestrial irradiance can be calculated using a formula that accounts for geometric relationship between the Sun and the Earth. Normalizing with extraterrestrial irradiance results in the development of forecasting models utilizing the clearness index \citep{sanfilippo2016adaptive, akarslan2018novel}. Most of the traditional approaches to solar irradiance forecasting rely on Clearsky model, computing an intermediate stationary  form (called Clearsky Index or \cyr{\texttt{\cyr{\texttt{CSI}}}}), as they typically produce better forecasts compared to models using clearness index \citep{lauret2022forecasts}. 
Methods employing Clearsky models \citep{hoyos2022short, david2016probabilistic, frimane2022infinite, shan2022ensemble, gneiting2023probabilistic} seek to use solar irradiance in cloud-free (clear sky) conditions and subsequently enhance forecasts by incorporating observed meteorological data characterizing the sky state and content in gas and particles with a seasonal and daily variation. 
Traditional forecasting methodologies face critical limitations, including inaccurate Clearsky models, elevated computational expenses associated with Numerical Weather Prediction (\cyr{\texttt{NWP}}) systems, and a limited ability to adjust to changing local meteorological conditions \citep{gueymard2021solar, sharma2020review}.
In order to overcome these limitations, this study proposes a framework that is based on a Clearsky-Free, data-driven, methodology that employs Extreme Learning Machines (\cyr{\texttt{ELM}}). This methodology simplifies the forecasting process, enhances adaptability, and provides accurate, real-time predictions that are tailored to modern energy systems by explicitly modeling Global Horizontal Irradiance (\cyr{\texttt{GHI}}) with enriched periodic and probabilistic features.

\subsection{Challenges with Clearsky-Based Models and the \cyr{\texttt{\cyr{\texttt{CSI}}}}}
Clearsky models, such as \cyr{\texttt{McClear}} \citep{lefevre2013mcclear}, \cyr{\texttt{REST2}} \citep{gueymard2021solar} and \cyr{\texttt{SPARTA}} \citep{ruiz-arias_sparta_2023}, are widely used for calculating solar irradiance under cloud-free  conditions. These models provide essential baselines using celestial mechanics and interpreting atmospheric effects related to aerosols, water vapor and ozone, with broad applications in climate science, agriculture, solar energy, and air quality monitoring \citep{allen1998crop, voyant2017machine, sharma2020review}.
Despite their versatility, Clearsky-based models (\textit{i.e.,} using the Clearsky model as a stationarization method) exhibit significant limitations, particularly in real-time operational contexts.:
\begin{itemize}
    \item \textit{Synchronization Issues}: Mismatched measurements and model outputs, especially at low solar elevations when shading and solar mask can intervene, lead to inaccuracies and reduced reliability \citep{diagne2013review};
    \item \textit{Dependence on Atmospheric Inputs}: Parameters such as aerosol optical depth and water vapor content are widely available thanks to atmospheric composition model or reanalyses such as \cyr{\texttt{CAMS}} or MERRA2. However, the accuracy of these parameters can be poor in some regions leading to unreliable clear sky irradiance estimation citep{gueymard2021solar};
    \item \textit{Gaps in Coverage}: \cyr{\texttt{CSI}} derived from these models (as the ratio of \cyr{\texttt{GHI}} to Clearsky irradiance), is undefined during nighttime and prone to significant errors during transitional periods such as dawn and dusk \citep{jiang2021comparison, voyant2017machine};
    \item \textit{Limited Adaptability}: Clearsky models struggle to represent diffuse irradiance at low solar elevations and to adapt to localized, dynamic weather conditions, particularly in complex terrains \citep{perez2017predictive};
    \item \textit{Surface Orientation Biases}: For forecasting \cyr{\texttt{PV}} power or solar radiation on tilted surface, the irradiance provided by Clearsky model needs to be projected on a inclined plane using so-called transposition model. The limited knowledge of the anisotropy of the diffuse radiance can introduce a bias that propagate through the forecasting pipeline \citep{blanc2019short}. 
\end{itemize}
For photovoltaic systems, these limitations are further compounded by dynamic factors such as inverter efficiency, maintenance schedules, and panel degradation, which Clearsky models cannot capture \citep{wen2022data}. As a result, these models, like they are actually used in prediction, could add complexity without significantly improving the accuracy of forecasts. The limitation of Clearsky-Based models underscore the need for alternative approaches where possible. New Clearsky-Free methodologies should demonstrate that it is possible to bypass clearsky model and its limitations by directly leveraging raw irradiance data (e.g., \cyr{\texttt{GHI}}, GTI, BNI, DHI) without relying on theoretical Clearsky baselines \citep{RUIZARIAS201810}. An explanation of the stationary process in solar time series prediction is provided in \ref{Statio}, emphasizing that this approach captures the stochastic nature of solar irradiance, enhances robustness and adaptability, and eliminates pre-processing and post-processing steps such as \cyr{\texttt{CSI}} normalization and rescaling, thereby simplifying the forecasting process. These Clearsky-Free methods, powered by advanced machine learning techniques like \cyr{\texttt{ELM}}, represent a significant shift in solar irradiance forecasting. They overcome the key limitations of traditional models for time-series based short-term forecast, providing scalable, accurate, and efficient solutions tailored to modern energy systems \citep{wen2022data, voyant2017machine}.

\subsection{Opportunities for Clearsky-Free Models}
The limitations of \cyr{\texttt{CSI}}-based forecasting methods highlight the need for alternatives that bypass Clearsky models. These Clearsky-independent approaches predict solar irradiance variables directly, without relying on theoretical baselines offers significant advantages with regard to real-time forecasting.
Clearsky-Free models avoid the need for atmospheric inputs that are often subject to errors, such as aerosol optical depth and related synchronization issues. Instead, these models rely on endogenous information, such as lagged values of \cyr{\texttt{GHI}}, in order to simplify the forecasting task and avoid cascading errors \citep{Danny2016}. They provide continuity for every hour of the day, even at the critical low-light sunrise and sunset hours, therefore improving grid management-energy storage \citep{diagne2013review}. The hypothesis explored in this paper is that some adaptative statistical model can learn implicitely from past observations the effects of aerosol, ozone and water vapour on the solar irradiance,  making the use of a clearsky model unnecessary. Correctly learning this component of atmospheric extinction is not a trivial task, since it is strongly modulated by the effect of clouds, which is several orders of magnitude greater.Not all statistical forecast approaches are capable of learning reliably the effect of aerosol, ozone and water vapour in varying weather conditions. We expect that advanced machine learning techniques, including Extreme Learning Machines, are able to capture complex temporal dependencies scalably and efficiently. Probabilistic extensions \citep{martin2021probabilistic} enable quantification of uncertainty, allowing for superior operational decision-making \citep{huang2006extreme}. Table~\ref{tab:comparative_analysis} provides an overview of the key differences between Clearsky-Based and Clearsky-Free approaches. This comparison underscores how Clearsky-Free methods tackle important challenges, making them especially effective for applications that demand adaptability, scalability, and efficiency. Moreover, such a Clearsky-Free approach have additional benefits: it would be less computationally complex, making them more robust and adaptive. 
\begin{table}
\centering
\caption{Comparison of Clearsky-Based Models and Clearsky-Free Models}
\label{tab:comparative_analysis}
\footnotesize
\begin{tabular}{|>{\centering\arraybackslash}p{4.5cm}|>{\centering\arraybackslash}p{6.0cm}|>{\centering\arraybackslash}p{6.5cm}|}
\hline
\textbf{Aspect} & \textbf{Clearsky-Based Models} & \textbf{Clearsky-Free Models} \\ \hline
\multirow{2}{*}{Atmospheric inputs Dependence} 
& High & Low \\ 
& aerosols, water vapor and ozone data & uses lagged \cyr{\texttt{GHI}} \\ \hline
\multirow{2}{*}{Applicability in real-time} 
& Limited & High \\ 
& synchronization and pre-processing & directly processes raw data \\ \hline
\multirow{2}{*}{Computational complexity} 
& High & Low \\ 
& advanced numerical models & simplified data-driven models \\ \hline
\multirow{2}{*}{Adaptability to local dynamics} 
& Limited & High \\ 
& struggles with localized weather patterns & handles localized and nonlinear variations \\ \hline
\multirow{2}{*}{Accuracy of forecasts} 
& Moderate & High \\ 
& sensitive to errors in atmospheric inputs & leverages endogenous, real-time data \\ \hline
\multirow{2}{*}{Infrastructure requirements} 
& High & Low \\ 
& powerful computing, robust data pipelines & works with simpler configurations \\ \hline
\multirow{2}{*}{Robustness to data gaps} 
& Low & High \\ 
& errors propagate in case of missing inputs & less affected by incomplete datasets \\ \hline
\multirow{2}{*}{Optimal use case} 
& For theoretical studies & For dynamic and data-scarce environments \\ 
& or well-instrumented sites & \\ \hline
\end{tabular}
\end{table}

\subsection{Objective of the Study}
Despite their utility, Clearsky-Based models face critical limitations in real-time forecasting. These include inaccuracies in atmospheric inputs, reliance on complex pre-processing steps, and limited adaptability in dynamic or data-sparse environments. Addressing these challenges requires an approach that is simpler, more robust, and less dependent on extensive datasets. The objectives of this study are:
\begin{itemize}
    \item Develop a Clearsky-Free forecasting framework that eliminates the need for Clearsky normalization and minimizes reliance on atmospheric data.
    \item Leverage historical and endogenous features, such as lagged Global Horizontal Irradiance (\cyr{\texttt{GHI}}) values, to simplify and streamline the forecasting process.
    \item Design a solution that is adaptable and capable of performing effectively in environments with sparse or incomplete data.
\end{itemize}
The proposed framework will be evaluated on three criteria: accuracy, to ensure precise solar irradiance forecasts; robustness, to deliver reliable performance across varying climatic conditions and geographic regions; and parsimony, to achieve high forecasting quality with minimal data and low computational complexity.
By addressing the shortcomings of traditional Clearsky-Based models, this study positions Clearsky-Free approaches as practical, scalable, and efficient solutions for modern energy systems. The emphasis on simplicity, adaptability, and data efficiency makes this framework particularly suitable for both data-rich and resource-constrained environments. 
The structure of this paper is as follows: Section \ref{sec:methodology} describes the methodology, including data collection, quality control, and the modeling framework. Section \ref{sec:Result_disc} presents and discusses the results, covering both deterministic and probabilistic forecasting approaches, as well as comparisons to benchmark models. Finally, Section \ref{sec:concl} provides the conclusions and outlines future research directions.

\section{Methodology}
\label{sec:methodology}
To effectively address the limitations of Clearsky-Based methods and leverage the potential of Clearsky-Free models, a robust methodology is required. This section details the approach adopted, beginning with the collection and preprocessing of high-resolution data to ensure the accuracy and reliability of the forecasting methodology.

\subsection{Data Collection and Quality Control (QC)}
\cyr{This study employs a univariate and endogenous forecasting approach, where Global Horizontal Irradiance (\cyr{\texttt{GHI}}) predicts itself. The dataset is sourced from the \cyr{\texttt{SIAR}} network of agroclimatic weather stations in Spain \citep{DESPOTOVIC2024123215}. It includes high-resolution \cyr{\texttt{GHI}} records from 76 stations over a four-year period, with a 30-min temporal resolution. This dataset offers the fine granularity required for short-term forecasting while encompassing a diverse range of climatic zones, from arid regions to highly variable coastal and mountainous areas, ensuring broad applicability \citep{carrasco2017agroclimatic}.
To ensure data reliability, a strict quality control process proposed by \citet{app12178529} was applied. This included detecting and rectifying anomalies such as sensor malfunctions, extreme weather impacts, or inconsistent measurements. The \cyr{\texttt{McClear}} model, recognized for its accurate estimation of clear-sky irradiance, was used as a reference for validating the \cyr{\texttt{GHI}} data \citep{lefevre2013mcclear}. Since \cyr{\texttt{McClear}} operates in hindcast mode (where required atmospheric inputs are only available two days after the observation date), it provides highly reliable theoretical clear-sky estimates, making it a robust basis for identifying and correcting errors in the historical dataset. However, this operational constraint limits its use in real-time forecasting.
Special attention was paid to sunrise and sunset periods, where low solar elevation angles are known to introduce higher uncertainty \citep{cervantes2019evaluation}. To account for temporal dependencies in \cyr{\texttt{GHI}}, lagged input variables were created, enabling dynamic adaptation to the inherent variability of solar irradiance \citep{yang2018review}. The dataset was divided into three years for training and validation and one year for testing, ensuring robust evaluation of the model’s generalization capabilities under both stable and dynamic weather scenarios.
In this context, the \cyr{\texttt{ELM}} model was trained using a carefully selected set of input features. The primary predictor is historical \cyr{\texttt{GHI}}, using past values to forecast future irradiance. Additionally, the clear-sky index $k_t$, defined as the ratio between the measured \cyr{\texttt{GHI}} and its theoretical counterpart from \cyr{\texttt{McClear}}, was included to provide an implicit reference for atmospheric attenuation effects. Temporal features, such as the hour of the day, day of the year, and seasonal indicators, were incorporated to capture diurnal and seasonal variations. Although no direct aerosol measurements are included, their impact on solar radiation is implicitly accounted for through the clear-sky \cyr{\texttt{GHI}} from \cyr{\texttt{McClear}}, which integrates aerosol optical properties from \cyr{\texttt{CAMS}} reanalysis data.
By structuring models around historical \cyr{\texttt{GHI}} and clear-sky references, they captures complex dependencies between atmospheric conditions and solar irradiance without requiring explicit \cyr{\texttt{AOD}} measurements. This approach simplifies the forecasting pipeline while maintaining high performance. Furthermore, it allows for a reliable evaluation of the necessity of clear-sky in forecasting accuracy.}

\subsection{Model Description} \label{sec:ModelDescription}
This study evaluates a diverse set of models for solar irradiance forecasting, combining deterministic and probabilistic approaches to address the challenges of traditional Clearsky normalization. These models are designed to capture temporal dependencies and the inherent periodic variability of solar irradiance data.
Naive predictors serve as benchmarks and include:
\begin{itemize}
    \item Persistence \cyr{\texttt{(P)}}: Assumes future \cyr{\texttt{GHI}} is equal to the most recent observed value \citep{yang2018review};
    \item Clearsky Model \cyr{\texttt{(CS)}}: Uses theoretical Clearsky irradiance while ignoring cloud attenuatation \cite{lefevre2013mcclear}; 
    \item Smart Persistence \cyr{\texttt{(SP)}}: Combines persistence with deviations from theoretical Clearsky irradiance to account for diurnal patterns \citep{diagne2013review};
    \item Climatology-Persistence \cyr{\texttt{CLIPER}}: Blends climatological averages with persistence to exploit autocorrelation \citep{YANG2019981};
    \item Exponential Smoothing (\cyr{\texttt{ES}}): Applies a weighted average to recent deviations from Clearsky irradiance \citep{makridakis1998forecasting};
    \item Autoregressive AR(2)-like (\cyr{\texttt{ARTU}}): Incorporates temporal dependencies and accounts for uncertainty \citep{voyant2017benchmarks};
    \item Combination Model (\cyr{\texttt{COMB}}): Aggregates \cyr{\texttt{SP}}, \cyr{\texttt{CLIPER}}, \cyr{\texttt{ES}}, and \cyr{\texttt{ARTU}} using ensemble methods \citep{voyant2017benchmarks}.
\end{itemize}
The reference model in the deterministic case is an autoregressive \cyr{\texttt{AR(p)}} model computed with classical least square optimisation. In this model, the \cyr{\texttt{GHI}} at time $ y_t $ is expressed as a linear combination of its past values: $ y_t = \phi_1 y_{t-1} + \phi_2 y_{t-2} + \dots + \phi_p y_{t-p} + \epsilon_t $, where $ \phi_i $ are the coefficients, $ p $ is the model order, and $ \epsilon_t $ represents the error term or residual. The coefficients $\phi_i $ are estimated by minimizing the sum of squared residuals $ \min_{\phi} \sum_{t=1}^T \left( y_t - \sum_{i=1}^p \phi_i y_{t-i} \right)^2 $. This \cyr{\texttt{AR}} model serves as a benchmark for assessing the performance of machine learning-based approaches in forecasting. It will be used in two versions: \cyr{\texttt{AR}} (without clear sky) and \cyr{\texttt{rAR}} (with clear sky). See Sections \ref{311} and \ref{312} for more details.
 The Extreme Learning Machine (\cyr{\texttt{ELM}} or \cyr{\texttt{EL}}) model is the central predictive framework in this study, providing a scalable and computationally efficient solution. It relies on a single-hidden-layer neural network architecture with sigmoid and Gaussian activation functions to capture both linear and non-linear relationships in the data. The model configuration includes:
\begin{itemize}
    \item Input and hidden layer optimization: For each forecast horizon, the number of input neurons and hidden neurons is adjusted from in-sample data (the best \cyr{\texttt{nRMSE}} defines the best configuration). For one particular site with high variability for example, the 1-hour horizon uses in average 78 inputs and 472 hidden neurons, while the 5-hour horizon uses 138 inputs and 450 hidden neurons. The hidden layer size is scaled approximately four times the input size for robust performance \citep{huang2006extreme}.
    \item Training process: Multiple random initializations are used to ensure robustness. Output weights are optimized using Ridge regression, $\beta = (H^\top H + \lambda I)^{-1} H^\top Y$, where $H$ is the hidden layer output matrix, $\lambda$ is the regularization parameter, and $Y$ is the target vector \citep{lukasik2020ridge}.
    \item Model selection: The configuration with the lowest normalized root mean square error (\cyr{\texttt{nRMSE}}) on the validation set is selected for testing.
\end{itemize}
\cyr{A detailed \cyr{\texttt{ELM}} definition is given in \ref{annex:elm}. Note that \cyr{\texttt{ELM}} provides a significant advantage over deep learning models in terms of both training and inference speed. Unlike iterative backpropagation-based methods, \cyr{\texttt{ELM}} training is performed through a single matrix inversion step, resulting in high computational efficiency. Furthermore, due to its simple architecture consisting of a single hidden layer, real-time inference is enabled, with predictions being generated in a few seconds. In this study, It was verified that incoming data can be processed for a dataset of approximately 70 sites, with an execution time of less than 30 seconds, confirming the model’s suitability for operational solar forecasting applications where rapid response times are required. Unlike physics-based models that integrate explicit aerosol-related parameters, the ELM operates in a purely data-driven manner. It statistically extracts the most relevant relationships between past observations and future irradiance without requiring a precise estimation of atmospheric attenuation mechanisms. This means that the model indirectly captures the impact of aerosols, clouds, and other atmospheric phenomena based on historical patterns but does not explicitly differentiate their contributions.}

Quantile Regression (\cyr{\texttt{QR}}) is employed as a reference probabilistic forecasting method due to its robustness and established performance in uncertainty quantification. \cyr{\texttt{QR}} predicts conditional quantiles of the response variable, allowing for the generation of reliable prediction intervals without assumptions about the underlying data distribution. The method optimizes an asymmetric loss function defined as:
\begin{equation}
L_{\tau}(u) =
\begin{cases} 
\tau u & \text{if } u \geq 0 \\
(\tau - 1)u & \text{if } u < 0
\end{cases}
\end{equation}
where $u = y - \hat{y}$ is the residual and $\tau$ represents the quantile level. By solving this minimization problem using linear programming, \cyr{\texttt{QR}} constructs prediction intervals that align with target coverage levels \citep{koenker2005quantile}.
In addition to \cyr{\texttt{QR}}, a non-parametric methodology is used to derive prediction intervals directly from residuals of deterministic forecasts \citep{martin2021probabilistic}. Lookup tables are generated during the training phase, capturing empirical relationships between forecast errors and observed values \citep{10.1063/5.0128131}. This approach avoids predefined assumptions about error distributions, making it particularly robust in dynamic atmospheric conditions. More information are given in \ref{lookup}.
The methodology used during simulations offers several key advantages. It provides comprehensive benchmarking by incorporating both naive predictors and advanced models, enabling rigorous evaluation. The framework is capable of handling high-dimensional data and varying forecast horizons with minimal computational overhead. Probabilistic forecasting is robustly implemented through \cyr{\texttt{QR}}  alongside non-parametric approaches for enhanced flexibility.

\subsection{Evaluation Metrics}
\label{metrics}
The performance of the forecasting models is assessed using both deterministic and probabilistic metrics to capture accuracy, bias, and reliability. This comprehensive evaluation framework is critical for validating the efficacy of Clearsky-Free methodologies in solar irradiance forecasting.
For deterministic forecasting, the following metrics are employed:
\begin{itemize}
    \item Normalized Root Mean Square Error (\cyr{\texttt{nRMSE}}) quantifies the average deviation of predictions from observed values, with a focus on penalizing larger errors,  \texttt{nRMSE} $= {\text{E}[y]}^{-1} \cdot \sqrt{\frac{1}{N} \sum_{i=1}^N \left( y_i - \hat{y}_i \right)^2} $
where $ y_i $ represents the observed values, $ \hat{y}_i $ the predicted values, $ N $ the total number of observations and $ \text{E}[y] $ represents the expected value or mean of the random variable $y$ \citep{makridakis1998forecasting};
    \item Normalized Mean Absolute Error (\texttt{nMAE}) provides a robust measure of accuracy, less sensitive to outliers compared to \cyr{\texttt{nRMSE}}, \texttt{nMAE} $= \text{E}[y]^{-1} \cdot \frac{1}{N} \sum_{i=1}^N \left| y_i - \hat{y}_i \right|;$
    \item R-Squared (\texttt{R}$^2 $) measures the proportion of variance in the observed data explained by the model, providing an indicator of goodness-of-fit, \texttt{ R}$^2 = 1 - (\sum_{i=1}^N \left( y_i - \hat{y}_i \right)^2)/(\sum_{i=1}^N \left( y_i - \text{E}[y] \right)^2)$; 
    \item Normalized Mean Bias Error (\texttt{nMBE}) captures systematic bias in the forecasts \citep{cervantes2019evaluation}, \texttt{nMBE} $= \text{E}[y]^{-1} \cdot \frac{1}{N} \sum_{i=1}^N \left( y_i - \hat{y}_i \right).$ 
\end{itemize}
For probabilistic forecasting, the evaluation framework includes:
\begin{itemize}
    \item Continuous Ranked Probability Score (\cyr{\texttt{CRPS}}) compares the cumulative distribution functions (\cyr{\texttt{CDF}}s) of predicted and observed values, \texttt{CRPS}$(F, y) = \int_{-\infty}^{\infty} \left( F(z) - \mathbf{1}_{z \geq y} \right)^2 dz$, evaluating the entire predictive distribution, where $ F(z) $ is the forecast \cyr{\texttt{CDF}}, and $ y $ is the observed value. This study computes \cyr{\texttt{CRPS}} using centiles, offering finer granularity than quartiles for solar irradiance forecasts \citep{gneiting2007strictly}. The quantile-based \cyr{\texttt{CRPS}} method used during this study is detailed in \ref{CRPS}; 
    \item Prediction Interval Coverage Probability (\cyr{\texttt{PICP}}) evaluates the reliability of prediction intervals by measuring the percentage of observed values falling within these intervals, \text{\cyr{\texttt{PICP}}}$ = \frac{1}{N} \sum_{i=1}^{N} \mathbf{1}_{\underline{y_i} \leq y_i \leq \overline{y_i}}$, 
    where $ \underline{y_i} $ and $ \overline{y_i} $ are the lower and upper bounds of the prediction interval \citep{martin2021probabilistic};
    \item Mean Interval Length (MIL) assesses the sharpness of prediction intervals, providing an indicator of interval width, $\text{MIL} = \frac{1}{N} \sum_{i=1}^N \left( \overline{y_i} - \underline{y_i} \right);$
 \item Interval Score (IS) evaluates the quality of prediction intervals by balancing two factors: sharpness (ensuring the intervals are as narrow as possible, measured by the MIL) and coverage reliability (Penalizing predictions when observed values fall outside the prediction intervals). The Interval Score is defined as: $    \text{IS} = \text{MIL} + \frac{2}{\alpha} \sum_{i=1}^N \left[ 
        \mathbf{1}_{y_i < \underline{y_i}} \left( \underline{y_i} - y_i \right) + 
        \mathbf{1}_{y_i > \overline{y_i}} \left( y_i - \overline{y_i} \right) 
    \right],$
    where $ \alpha $ is the nominal coverage level of the prediction interval, and $ \mathbf{1} $ denotes the indicator function (equals 1 if the condition is true and 0 otherwise) \citep{gneiting2007strictly}.
\end{itemize}
The evaluation framework uses deterministic metrics like \cyr{\texttt{nRMSE}} and \cyr{\texttt{nMAE}}, along with probabilistic metrics like \cyr{\texttt{CRPS}} and \cyr{\texttt{PICP}}, to assess model reliability, accuracy, and practical relevance. The Mann-Whitney U test \citep{Mann-Whitney-U-test}, a non-parametric method, is used to compare distributions without assuming normality, making it ideal for scenarios where error distributions deviate from Gaussian norms. The test measures shifts in forecast error distributions, with a p-value below the standard threshold of 0.05 indicating a significant difference. For example, Model B outperforms Model A with a p-value of 0.021, indicating superior performance with consistently lower forecast errors.

\section{Results and Discussion}
\label{sec:Result_disc}
In this section, the results for both deterministic and probabilistic forecasting are presented and analyzed. The \cyr{\texttt{ELM}} configuration (size of input and hidden nodes), was optimized for each prediction horizon using the Nelder-Mead simplex algorithm \citep{nelder1965simplex}, with an average of 85 input nodes and 430 hidden nodes. The ridge parameter is set to $\lambda = 0.2$, and we arbitrarily select a 60\%-40\% hybrid transfer function mix of sigmoidal and Gaussian functions, respectively, to ensure both flexibility and efficiency. Training involved 96 runs (with in-sample data) to identify optimal weights of \cyr{\texttt{ELM}}. While simulations used a high-performance computing system (1840 CPU cores and 6.4 TB of RAM), \cyr{\texttt{ELM}} required less than a minute for one year of training and forecasting. The longer total time (under 10 min per site with the possibility to parallelize certain tasks) was mainly due to the computational demands of \cyr{\texttt{QR}}. Violin plots will be used to compare distributions across groups. They combine boxplots and kernel density plots, showing data spread, density, and central tendency. This allows for intuitive visualization of differences in shape, range, and median between groups.

\subsection{Deterministic Case}
Point prediction, or deterministic prediction, involves generating a single estimated value for a future variable without accounting for uncertainty or variability. This approach is crucial as it enables precise and timely decision-making in applications such as solar energy management, where accurate forecasts can optimize resource utilization and enhance system efficiency.

\subsubsection{Justifying the Use of \cyr{\texttt{ELM}}}
\label{311}
\begin{figure}
    \centering
    \begin{subfigure}[b]{0.45\textwidth}
        \centering
        \includegraphics[width=\textwidth]{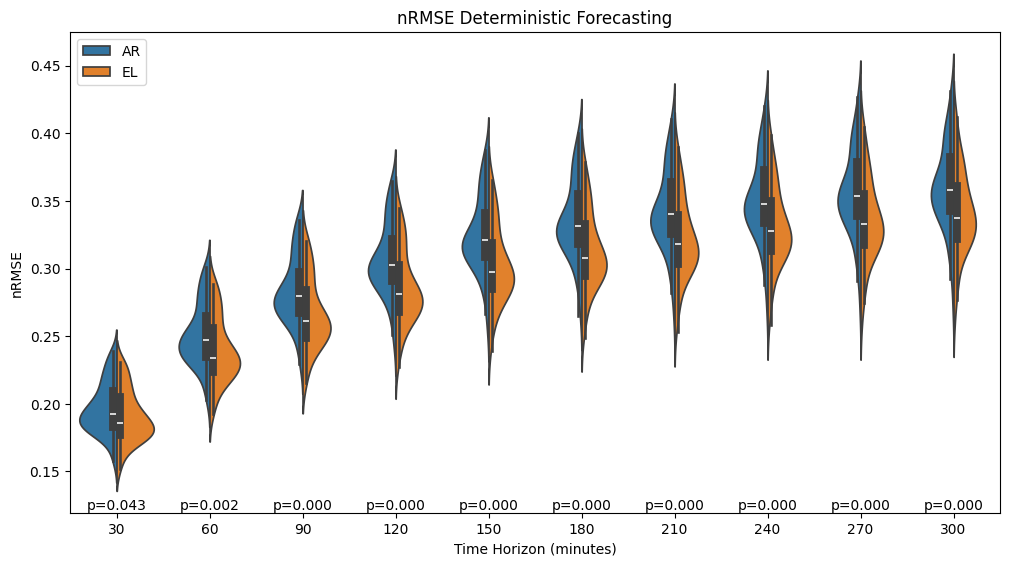}
        \caption{nRMSE}
        \label{fig:nRMSE_plot}
    \end{subfigure}
    \hfill
    \begin{subfigure}[b]{0.45\textwidth}
        \centering
        \includegraphics[width=\textwidth]{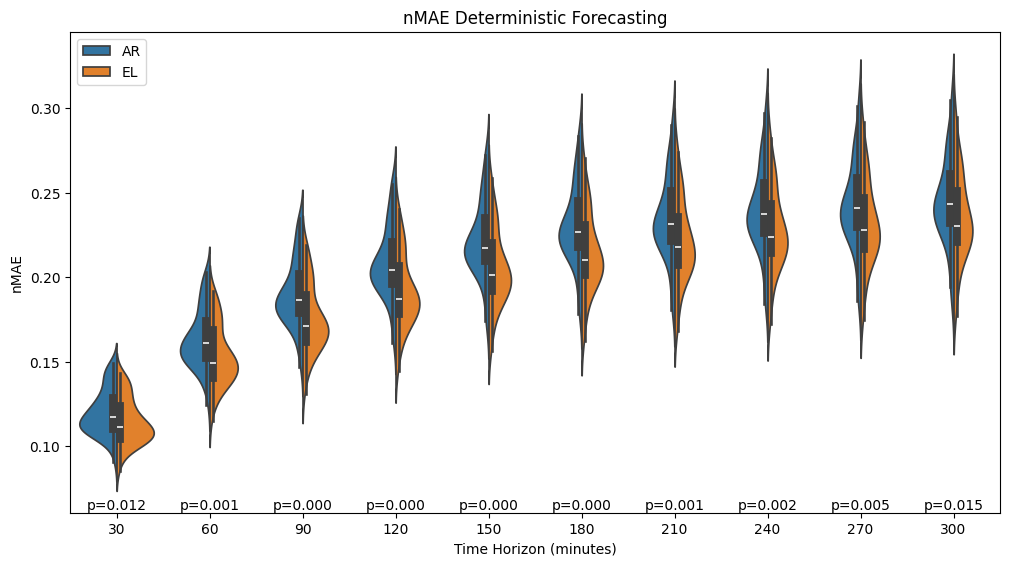}
        \caption{nMAE}
        \label{fig:nMAE_plot}
    \end{subfigure}

    \vspace{0.5cm}
    \begin{subfigure}[b]{0.45\textwidth}
        \centering
        \includegraphics[width=\textwidth]{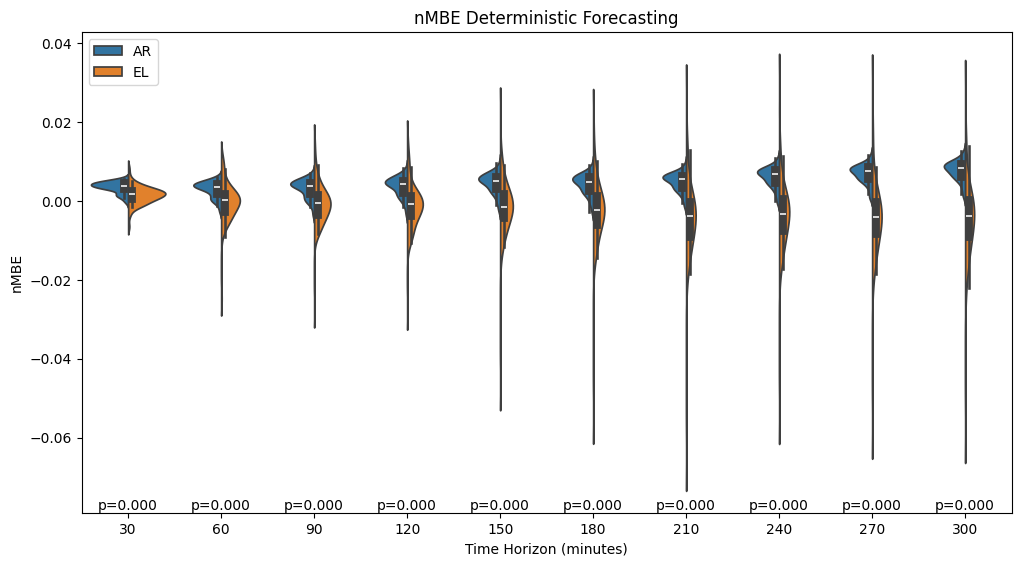}
        \caption{nMBE}
        \label{fig:nMBE_plot}
    \end{subfigure}
    \hfill
    \begin{subfigure}[b]{0.45\textwidth}
        \centering
        \includegraphics[width=\textwidth]{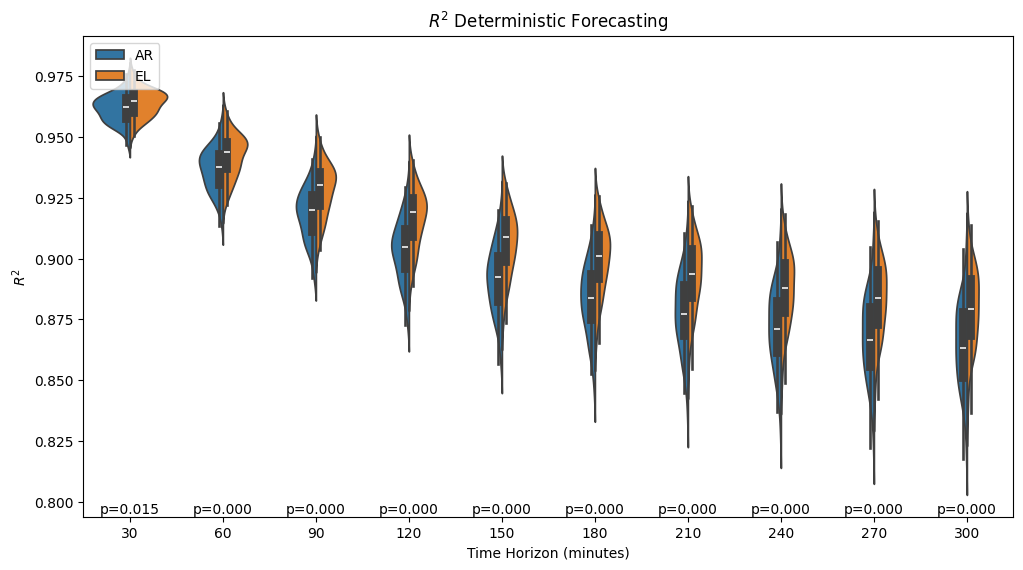}
        \caption{$R^2$}
        \label{fig:R2_plot}
    \end{subfigure}
    \caption{Evaluation metrics for deterministic forecasting over the 76 \cyr{\texttt{SIAR}} Stations. Each plot illustrates a specific metric (\cyr{\texttt{nRMSE}}, \cyr{\texttt{nMAE}}, \cyr{\texttt{nMBE}}, \cyr{\texttt{R²}}) across 10 time horizons (30 to 300 min). The comparison includes two models: \cyr{\texttt{AR}} (blue) and \cyr{\texttt{ELM}} (orange).}
    \label{fig:metrics_combined}
\end{figure}

Initially, it is essential to validate the use of \cyr{\texttt{ELM}}  against a purely linear autoregressive \cyr{\texttt{AR}}($p$) model, optimizing $p$ for each site and forecasting horizon using a brute force method comparing in-sample results for $p=1$ to $p=96$ ($p$ median $=58$). The comparison of performances of both models is operated on raw data (without employing Clearsky models). Note that $p$ can't be defined from Partial Autocorrelation Function (\cyr{\texttt{PACF}}) while this parameter is periodic with raw data. 
Performance of the deterministic forecasting, across various metrics and time horizons are presented in Figure \ref{fig:metrics_combined}. The resulting p-values from Mann-Whitney U test are displayed below the respective plots for each time horizon in the figures. 
The results show that \cyr{\texttt{ELM}} consistently outperforms \cyr{\texttt{AR}} across all error metrics (\cyr{\texttt{nRMSE}}, \cyr{\texttt{nMAE}}, \cyr{\texttt{nMBE}}, \cyr{\texttt{R²}}) and time horizons, with very low p-values confirming the significance of these differences. Therefore, \cyr{\texttt{ELM}} is clearly the better choice for more accurate, reliable, and unbiased predictions.

\subsubsection{Validation Against Reference and Naive Predictors}
\label{312}
\begin{figure}
    \centering
    \includegraphics[width=0.8\textwidth]{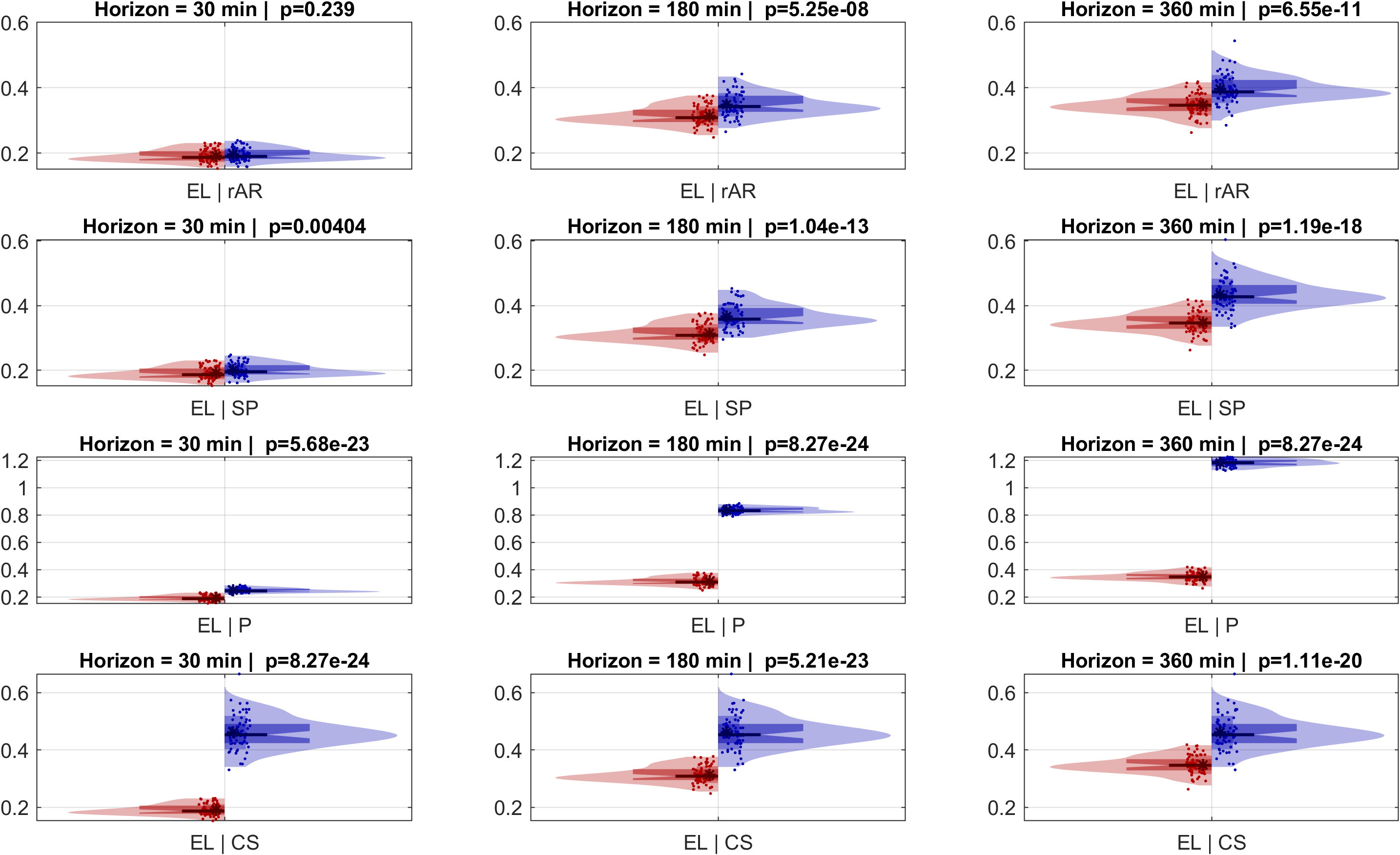}
    \caption{Comparison of the \cyr{\texttt{nRMSE}} for deterministic forecasting for time horizons 30 min, 180 min, and 360 min for the \cyr{\texttt{EL}} model (left) and four predictors (right): \cyr{\texttt{rAR, SP, P,}} and \cyr{\texttt{CS}}.}
    \label{fig:p_naive_Part1}
\end{figure}

To validate the \cyr{\texttt{ELM}} Clearsky-Free approach, predictions are compared to classical solar irradiance tools, which are defined as reference (\cyr{\texttt{rAR}}) and naive models (\cyr{\texttt{SP, P,}} and \cyr{\texttt{CS}}) derived from the \cyr{\texttt{McClear}} Clearsky model. The order $ p $ of the \cyr{\texttt{rAR}}($ p $) model is optimized using a \cyr{\texttt{PACF}}-based approach: the first minimum of the \cyr{\texttt{PACF}} function determines the maximum lag to consider, which corresponds to $ p $ ($p$ median $=6$).
As shown in Figure \ref{fig:p_naive_Part1}, \cyr{\texttt{ELM}}  consistently outperforms all naive models (\cyr{\texttt{SP, P,}} and \cyr{\texttt{CS}}) across all time horizons (30, 180, and 360 min) with highly significant differences (p < 0.001). While \cyr{\texttt{ELM}} also surpasses \cyr{\texttt{rAR}} at 180 and 360 min (p < 0.001), the difference is not significant for the 30-min horizon (p = 0.239). Among the models, \cyr{\texttt{EL}} ranks as the best, followed by \cyr{\texttt{rAR, SP, CS,}} and finally \cyr{\texttt{P}}. 

\subsubsection{Benchmark Expansion and \cyr{\texttt{ELM}}'s Dominance}
\begin{figure}
    \centering
    \includegraphics[width=0.8\textwidth]{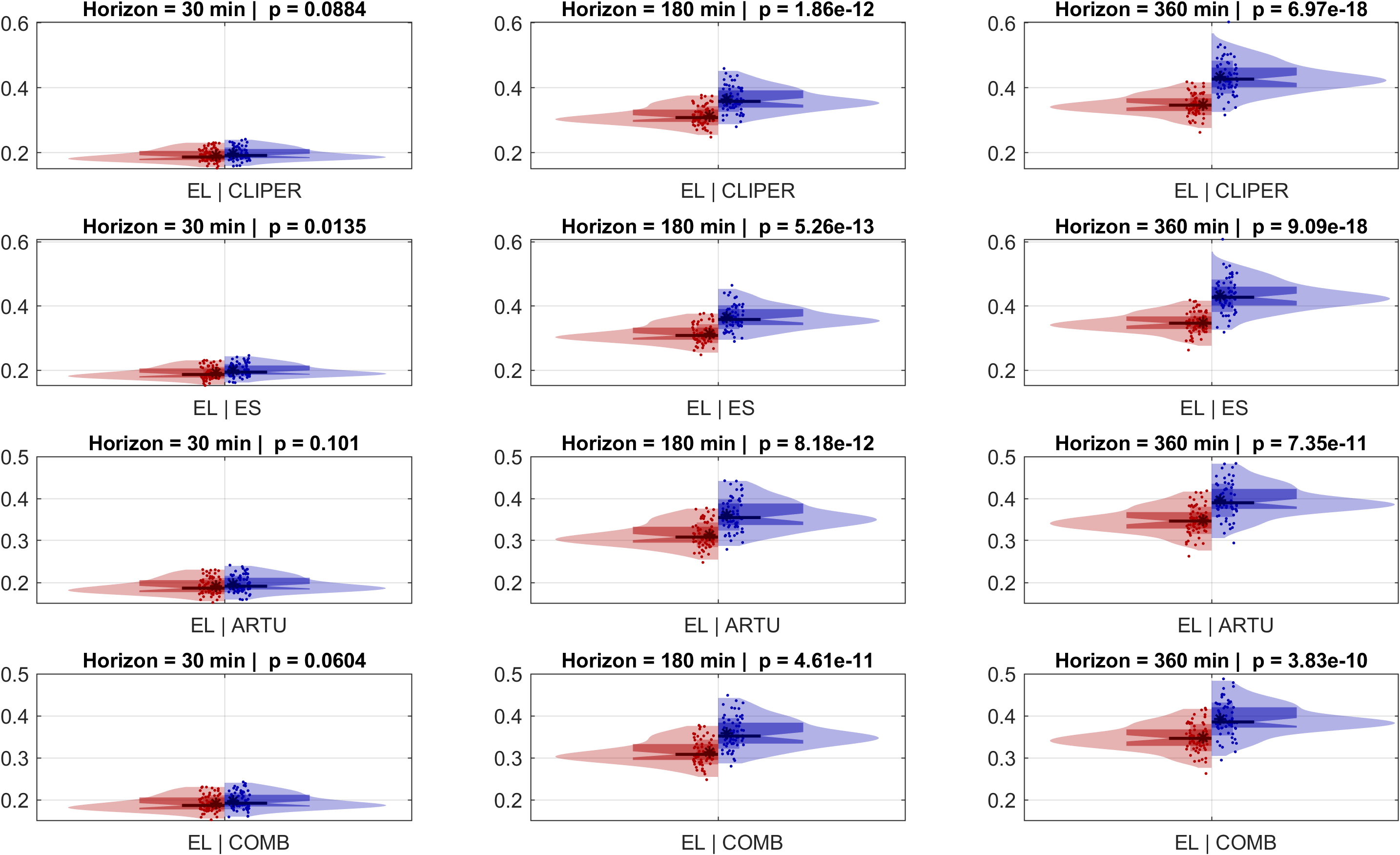}
    \caption{Comparison of the \cyr{\texttt{nRMSE}} for deterministic forecasting for time horizons 30 min, 180 min, and 360 min for the \cyr{\texttt{EL}} model (left) and four benchmark predictors (right): \cyr{\texttt{CLIPER, ES, ARTU,}} and \cyr{\texttt{COMB}}.} 
    \label{fig:p_naive_Part2}
\end{figure}

Expanding the benchmark with additional models strengthens the conclusions regarding \cyr{\texttt{ELM}}'s superiority (Figure \ref{fig:p_naive_Part2}). It consistently outperforms all benchmark models (\cyr{\texttt{CLIPER, ES, ARTU,}} and \cyr{\texttt{COMB}}) across medium (180 min) and long (360 min) horizons, with highly significant improvements ($p < 10^{-11}$). At the short 30-min horizon, the differences are less pronounced, with some comparisons showing non-significant p-values (e.g., \cyr{\texttt{CLIPER}} and \cyr{\texttt{ARTU}}), though \cyr{\texttt{EL}} still performs competitively. Overall, \cyr{\texttt{ELM}} is the most reliable and effective model, particularly excelling in medium and long-term forecasts.

\subsubsection{Ranking in Deterministic Case}
In conclusion, the \cyr{\texttt{ELM}} model demonstrates superior performance across all time horizons, particularly excelling in medium and long-term forecasts. While differences are less pronounced at the 30-min horizon, \cyr{\texttt{ELM}} remains competitive, solidifying its position as the most reliable and effective forecasting model. \ref{bench} provide additional information allowing to finally rank all the tested models in the case of point prediction. From best to worst is as follows: \cyr{\texttt{EL, AR, rAR, CLIPER, SP, ARTU, ES, COMB, CS}} and \cyr{\texttt{P}}. 
This validates our assumption that a well-chosen adaptative statistical model can efficiently learn the attenuation due to aerosol, water vapour and ozone from previous measurements. The comparison with simpler models further demonstrates that this can only be achieved using advanced statistical models. A comparison with approaches based on clearsky model finally shows that, when learned from past observations, cloud-free atmospheric modelling is more accurate that when this information is drawn from a Clearsky model. 

\subsection{Probabilistic Case}
Including probabilistic forecasts complements deterministic predictions by quantifying uncertainty, which is critical for optimizing energy systems. This approach allows better risk management, improved resource allocation, and enhanced decision-making under variability, leveraging advanced statistical methods to capture forecast distributions and tail risks.

\subsubsection{Median Estimation from Probabilistic Case}
\begin{figure}
    \centering
    \begin{subfigure}[b]{0.45\textwidth}
        \centering
        \includegraphics[width=\textwidth]{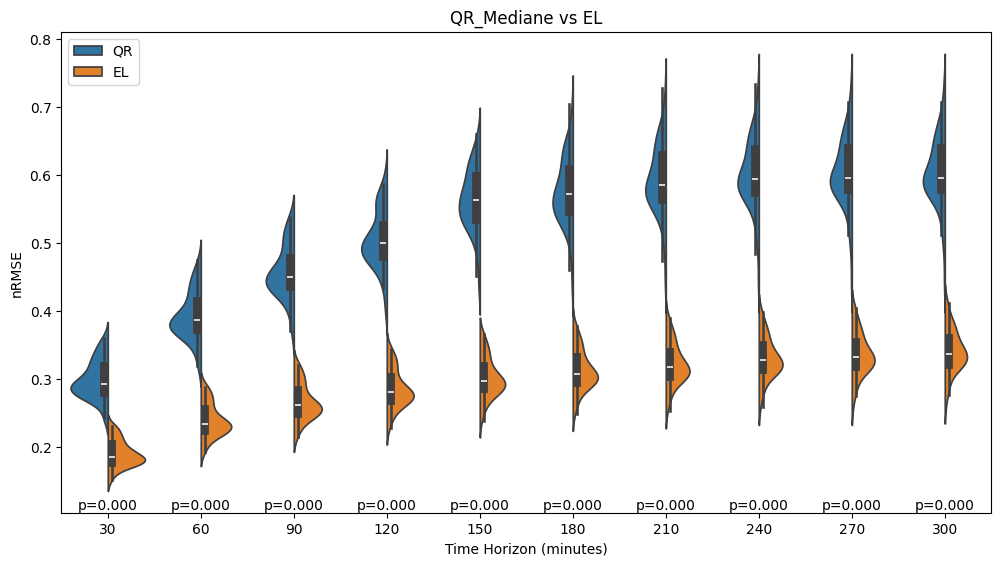}
        \caption{nRMSE}
        \label{fig:p_nRMSE_QR_Mediane_plot}
    \end{subfigure}
    \hfill
    \begin{subfigure}[b]{0.45\textwidth}
        \centering
        \includegraphics[width=\textwidth]{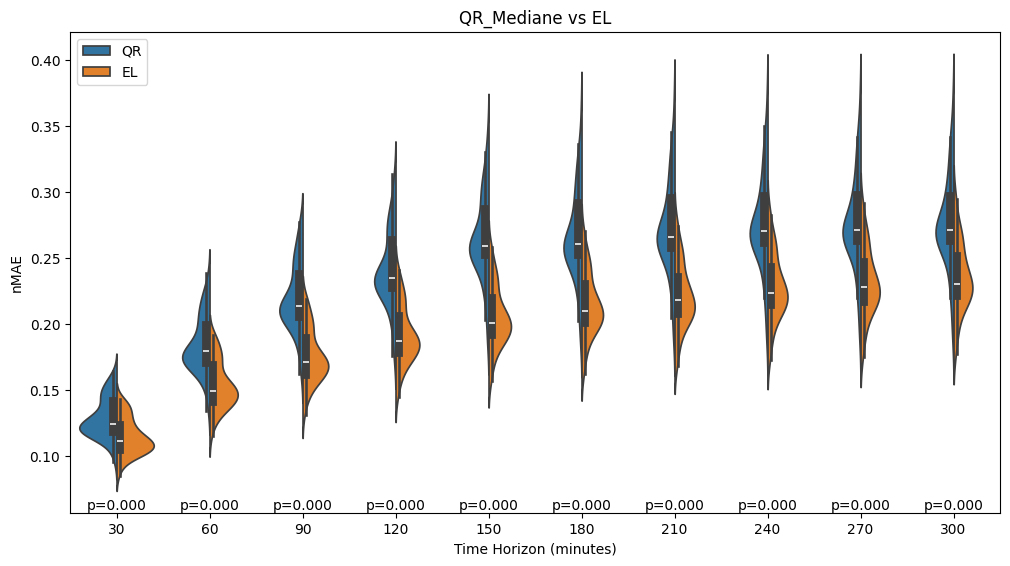}
        \caption{nMAE}
        \label{fig:p_nMAE_QR_Mediane_plot}
    \end{subfigure}
    \vspace{0.5cm}
    \begin{subfigure}[b]{0.45\textwidth}
        \centering
        \includegraphics[width=\textwidth]{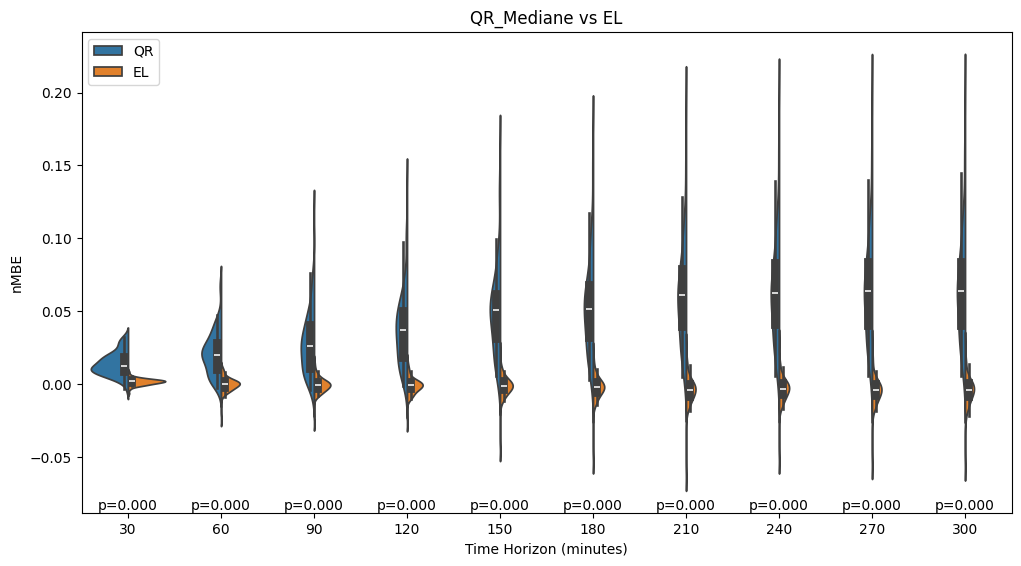}
        \caption{nMBE}
        \label{fig:p_nMBE_QR_Mediane_plot}
    \end{subfigure}
    \hfill
    \begin{subfigure}[b]{0.45\textwidth}
        \centering
        \includegraphics[width=\textwidth]{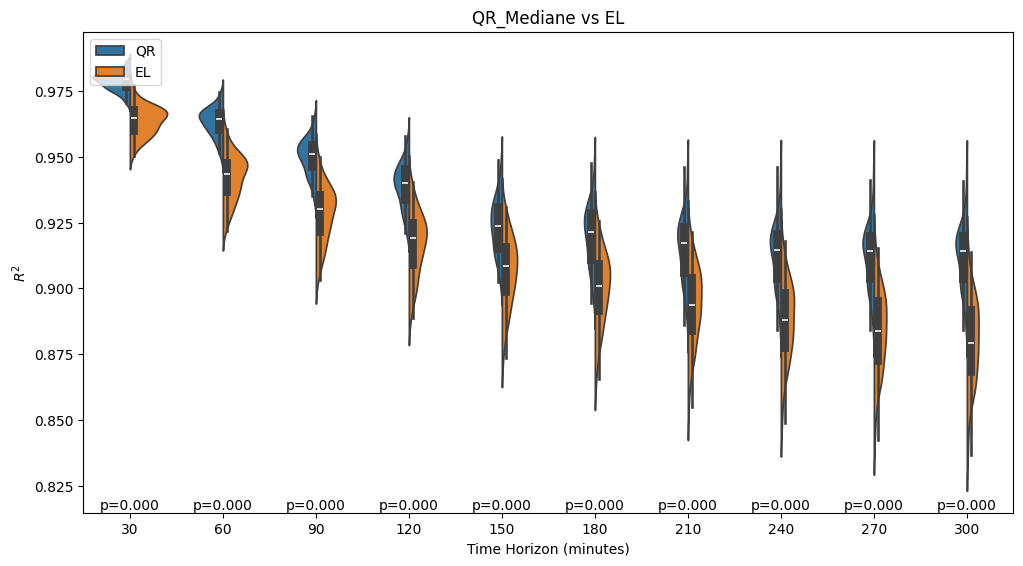}
        \caption{$R^2$}
        \label{fig:p_R2_QR_Mediane_plot}
    \end{subfigure}
    \caption{Evaluation metrics for deterministic forecasting over the 76 \cyr{\texttt{SIAR}} Stations. Each plot illustrates a specific metric (\cyr{\texttt{nRMSE}}, \cyr{\texttt{nMAE}}, \cyr{\texttt{nMBE}}, $R^2$) across 10 time horizons (30 to 300 min). The comparison includes two models: \cyr{\texttt{QR}} median (blue) and \cyr{\texttt{ELM}} (orange).}
    \label{fig:QR_median_combined}
\end{figure}

One advantage of working directly with probabilistic forecasts is that estimating the median, or 0.5 quantile, inherently provides a deterministic prediction. In the case of Quantile Regression (\cyr{\texttt{QR}}), this facilitates seamless integration of probabilistic and deterministic approaches. Similarly, for \cyr{\texttt{ELM}}, the methodology based on lookup tables ensures that both probabilistic and deterministic forecasts are generated concurrently, enhancing efficiency and flexibility. 
In this section, and particularly in Figure \ref{fig:QR_median_combined}, the results comparing the medians obtained from \cyr{\texttt{QR}} and \cyr{\texttt{ELM}} are presented. Overall, \cyr{\texttt{ELM}} outperforms \cyr{\texttt{QR}} Median across most metrics and time horizons, particularly in medium (180 min) and long-term (360 min) forecasts. For \cyr{\texttt{nRMSE}} and \cyr{\texttt{nMAE}}, \cyr{\texttt{ELM}} consistently achieves lower errors, indicating higher predictive accuracy at all horizons. In terms of \cyr{\texttt{nMBE}}, it demonstrates less bias than \cyr{\texttt{QR}}, remaining closer to zero across all horizons. For \cyr{\texttt{R²}}, \cyr{\texttt{QR}} shows a slight advantage at the 30-min horizon but declines significantly at longer horizons.

\subsubsection{\cyr{\texttt{QR}} Versus \cyr{\texttt{ELM}}}
\begin{figure}
    \centering
    \begin{subfigure}[b]{0.45\textwidth}
        \centering
        \includegraphics[width=\textwidth]{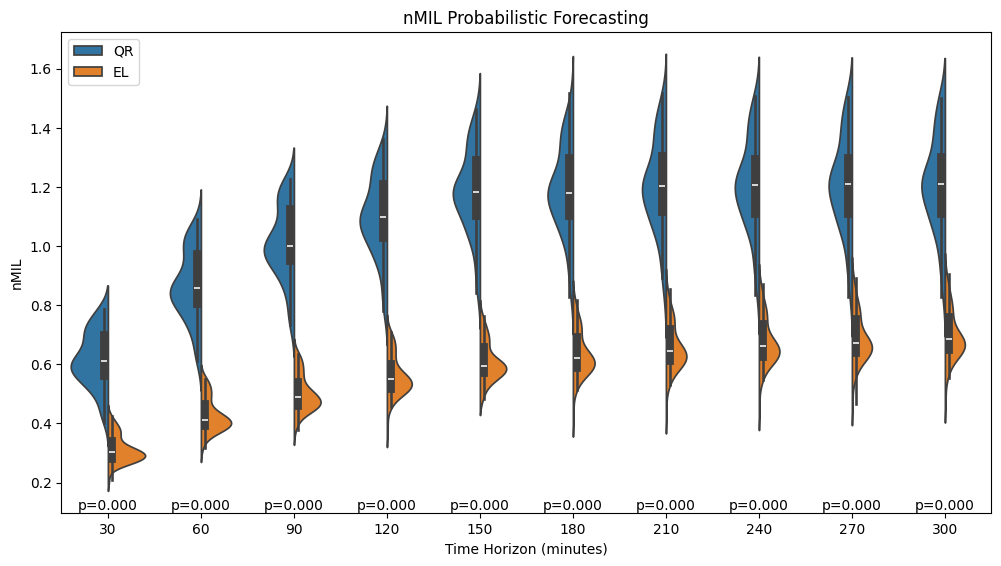}
        \caption{nMIL}
        \label{fig:p_nMIL_plot}
    \end{subfigure}
    \hfill
    \begin{subfigure}[b]{0.45\textwidth}
        \centering
        \includegraphics[width=\textwidth]{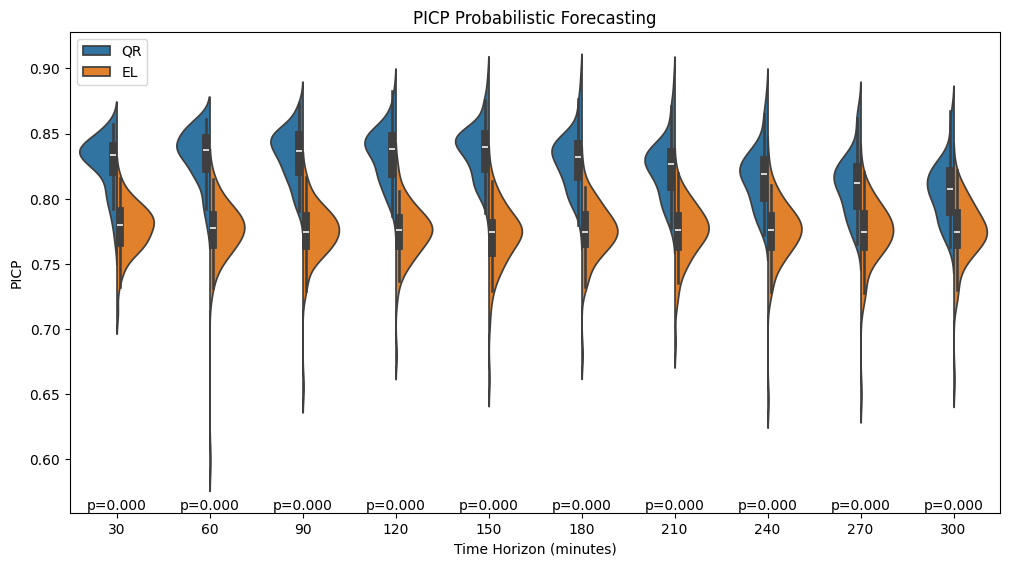}
        \caption{PICP}
        \label{fig:p_PICP_plot}
    \end{subfigure}
    \vspace{0.5cm}
    \begin{subfigure}[b]{0.45\textwidth}
        \centering
        \includegraphics[width=\textwidth]{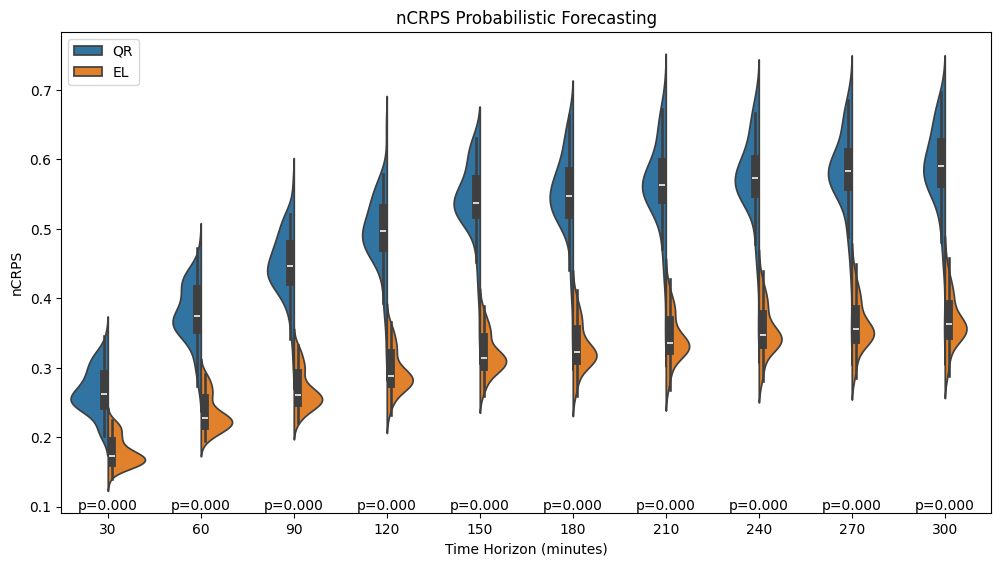}
        \caption{nCRPS}
        \label{fig:p_nCRPS_plot}
    \end{subfigure}
    \hfill
    \begin{subfigure}[b]{0.45\textwidth}
        \centering
        \includegraphics[width=\textwidth]{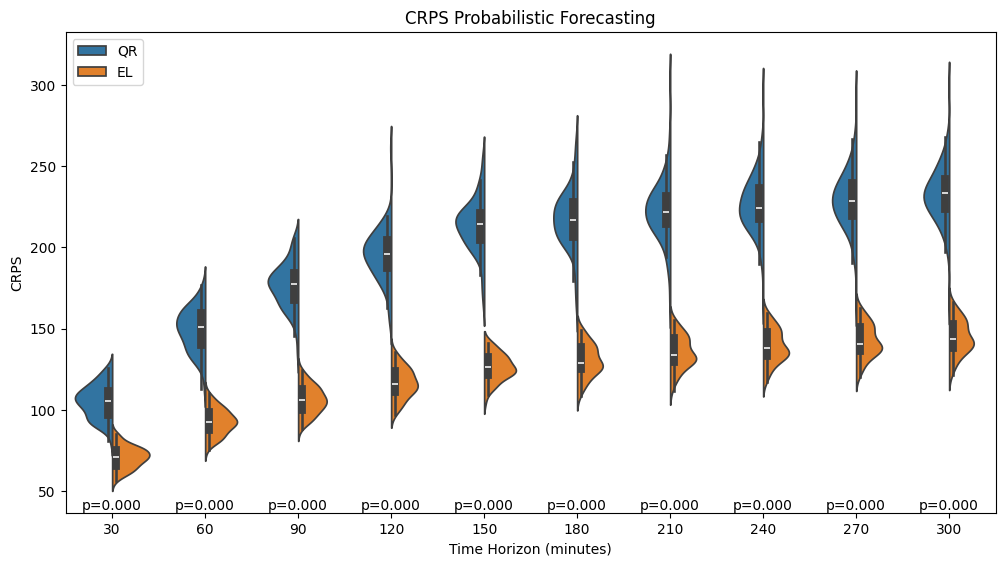}
        \caption{CRPS}
        \label{fig:p_CRPS_plot}
    \end{subfigure}
    \caption{Probabilistic forecasting metrics across 10 time horizons (30 to 300 min, in 30-min increments). Metrics include nMIL, \cyr{\texttt{PICP}}, \cyr{\texttt{nCRPS}}, and \cyr{\texttt{CRPS}} for two models: \cyr{\texttt{QR}} (blue) and EL (orange). Statistical p-values for comparison are provided below the violins for each time horizon.}
    \label{fig:probabilistic_metrics_combined}
\end{figure}

\begin{figure}
    \centering
    \includegraphics[width=0.5\textwidth]{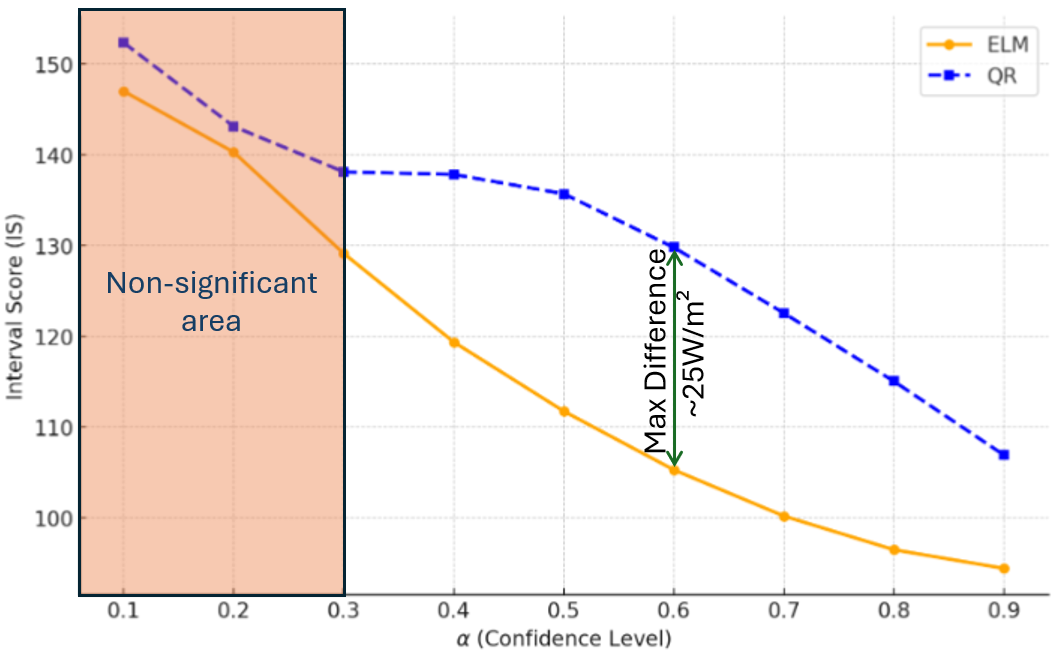} 
    \caption{Interval Score (IS) comparison for 30-min horizon and for \cyr{\texttt{ELM}} and \cyr{\texttt{QR}} models across confidence levels $\alpha$.}
    \label{fig:IS_comparison}
\end{figure}

Probabilistic error metrics across ten different forecast horizons are presented in Figure \ref{fig:probabilistic_metrics_combined}. The \cyr{\texttt{nMIL }}metric (Figure \ref{fig:p_nMIL_plot}) evaluates the average width of the prediction intervals, where lower values indicate narrower intervals and thus higher confidence in the forecasts. The results show that both \cyr{\texttt{QR}} and \cyr{\texttt{EL}} maintain consistent interval widths across horizons of 150 min and above, while \cyr{\texttt{EL}} generally provides tighter intervals compared to \cyr{\texttt{QR}}, particularly at longer horizons, indicating higher precision in \cyr{\texttt{EL}} probabilistic forecasts. 
The \cyr{\texttt{PICP}} metric (Figure \ref{fig:p_PICP_plot}) reflects the percentage of actual observations captured within the predicted intervals. For this study, the nominal value is set at $\alpha = 0.2$, corresponding to an 80\% prediction interval. A well-calibrated model should have \cyr{\texttt{PICP}} values close to 80\%. The results indicate that the \cyr{\texttt{QR}} model consistently achieves \cyr{\texttt{PICP}} values closer to 80\%, suggesting better calibration. In contrast, the \cyr{\texttt{EL}} model's slightly lower \cyr{\texttt{PICP}} values highlight a trade-off, as its narrower intervals prioritize precision but might compromise calibration in terms of coverage reliability.
The \cyr{\texttt{CRPS}} metric (Figure \ref{fig:p_CRPS_plot}) evaluates the alignment between forecast distributions and actual observations, where lower values indicate better performance. The \cyr{\texttt{EL}} model consistently outperforms \cyr{\texttt{QR}} across all horizons, indicating that its forecast distributions are closer to actual observations. Similarly, the \cyr{\texttt{nCRPS}} metric (Figure \ref{fig:p_nCRPS_plot}), which normalizes \cyr{\texttt{CRPS}} values, reinforces \cyr{\texttt{EL}} superior performance by highlighting its ability to provide accurate and consistent probabilistic forecasts relative to observed data.
For \cyr{\texttt{GHI}} forecasting over a 30-min horizon (mean concerning all sites), the mean Interval Scores (\cyr{\texttt{IS}}) were 115.9 W/m² for \cyr{\texttt{ELM}} and 131.2 W/m² for \cyr{\texttt{QR}}, as shown in Figure \ref{fig:IS_comparison}. The lower mean \cyr{\texttt{IS}} indicates that \cyr{\texttt{ELM}} is the better model for this horizon. A significance test comparing the \cyr{\texttt{IS}} distributions yielded a p-value of 0.075 for $\alpha < 0.3$, indicating that the difference between \cyr{\texttt{ELM}} and \cyr{\texttt{QR}} is not statistically significant at higher confidence levels ($ \geq 70\%$). However, as $\alpha$ increases, the p-value decreases significantly, suggesting that the performance gap becomes statistically significant at lower reliability levels ($ < 70\%$). 
Similar patterns are observed for other forecast horizons, where \cyr{\texttt{ELM}} consistently achieves lower mean IS values than \cyr{\texttt{QR}}. These results, reflected in comparable interval score curves across horizons, confirm that \cyr{\texttt{ELM}} provides superior probabilistic forecasting performance. The statistical significance of the differences, however, varies depending on the reliability level and forecast horizon.

\subsubsection{Ranking in Probabilistic Case}
\begin{figure}
    \centering
    \includegraphics[width=0.8\textwidth]{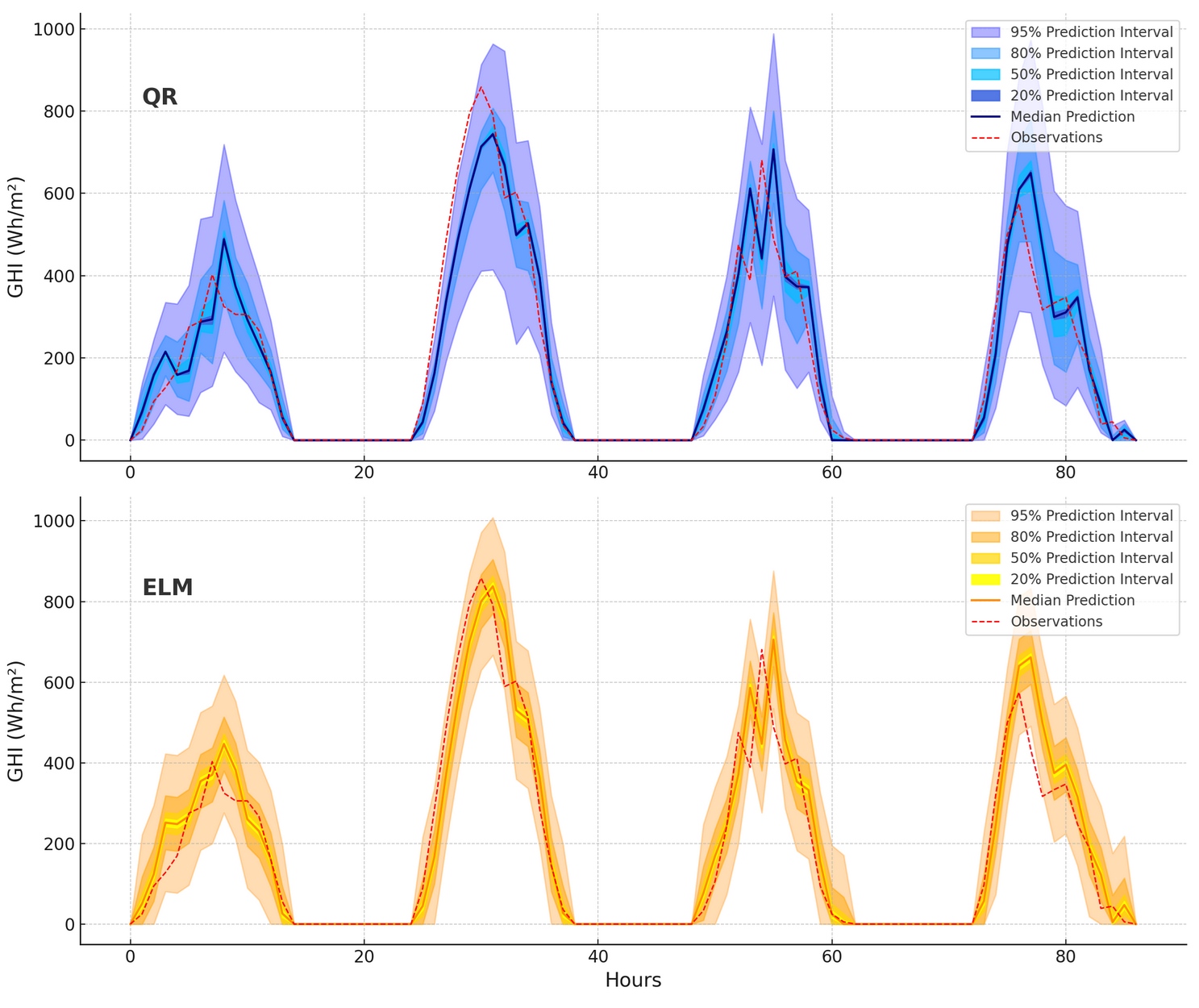}
    \caption{Prediction intervals and median forecasts for Quantile Regression (\cyr{\texttt{QR}}, top) and Extreme Learning Machine (\cyr{\texttt{ELM}}, bottom) models for \cyr{\texttt{GHI}} over an extended period of four days and an horizon 1h.}
    \label{fig:prob_predictions}
\end{figure}

The comparison of prediction intervals for \cyr{\texttt{QR}} and \cyr{\texttt{ELM}} models (Figure~\ref{fig:prob_predictions}) highlights the superior performance of \cyr{\texttt{ELM}} in probabilistic forecasting. \cyr{\texttt{ELM}} provides consistently narrower prediction intervals across all confidence levels (95\%, 80\%, 50\%, and 20\%), demonstrating reduced uncertainty and more confident predictions. In contrast, \cyr{\texttt{QR}} exhibits significantly wider intervals, particularly at upper than 95\% confidence level, indicating higher prediction variance and less precise forecasts. 
Additionally, the median forecasts (solid lines) from \cyr{\texttt{ELM}} align more closely with the observed \cyr{\texttt{GHI}} values (dotted red line), showcasing better calibration and accuracy compared to \cyr{\texttt{QR}}. While \cyr{\texttt{QR}} captures a larger range of possible outcomes due to its wider intervals, this comes at the cost of reduced precision and potentially less actionable predictions. 
This trade-off between precision and coverage further supports the selection of \cyr{\texttt{ELM}} for applications requiring high-confidence, actionable forecasts in energy systems.

\subsection{Synergies Between Clearsky and Clearsky-Free Approaches}
The Clearsky-Free approach demonstrated in this study provides robust results in both deterministic and probabilistic forecasting, but there is substantial potential to further enhance accuracy by integrating the strengths of Clearsky models such as \cyr{\texttt{McClear}} and its forecast-oriented counterpart, \cyr{\texttt{CAMS}}. These models provide high-quality irradiance baselines, while \cyr{\texttt{ELM}} excel at dynamically adapting to local atmospheric variability and temporal dependencies. By coupling these approaches, a hybrid model could outperform traditional methods, leveraging the theoretical precision of Clearsky models and the adaptability of \cyr{\texttt{ELM}}.
Clearsky models can also enhance the reliability of input data through quality control, improving the overall robustness of the forecasting system. This is particularly beneficial in regions with sparse or unreliable ground-based measurements, where satellite-derived models like \cyr{\texttt{McClear}} serve as dependable references.
A promising future direction involves creating hybrid models that combine the theoretical baselines of \cyr{\texttt{McClear}} with the adaptability of \cyr{\texttt{ELM}} for real-time atmospheric variations. Such models would improve solar irradiance forecasts, particularly in regions with complex dynamics. To enhance robustness and generalizability, training datasets should include diverse climatic conditions, and transfer learning techniques should be integrated to enable global applicability without location-specific historical data. These advancements would provide accurate predictions for any irradiance component (e.g., \cyr{\texttt{GHI}}, DHI, BNI, GTI) and configurations like tilted or azimuthal orientations.
Another key research avenue is refining probabilistic forecasting through advanced uncertainty quantification techniques, such as Gaussian Processes or Bayesian Neural Networks, to improve forecast reliability. Additionally, the application of Clearsky-Free methodologies in decision-support systems for energy grid operators must be explored, ensuring scalability and computational efficiency for real-time operational contexts.
By pursuing these directions, Clearsky-Free methodologies can evolve into more reliable and flexible tools for solar irradiance forecasting, addressing diverse operational and environmental challenges.

\section{Conclusion}
\label{sec:concl}
This article presents a groundbreaking method for solar irradiance forecasting without relying on the Clearsky model, offering performance that rivals or surpasses traditional approaches dependent on Clearsky. Specifically, an adaptive statistical model effectively learns the attenuation caused by aerosols, water vapor, and ozone using past measurements. By eliminating intermediate normalization steps, the methodology streamlines the forecasting process and improves reliability, paving the way for more accurate and efficient atmospheric modeling. Leveraging raw \cyr{\texttt{GHI}} data and advanced machine learning techniques such as Extreme Learning Machines, it captures solar irradiance's periodicity and local variability, with deterministic and probabilistic forecasting providing robust uncertainty estimates.
Conventional Clearsky-Based models are limited by synchronization issues, inaccuracies at night, and uncertainties at low solar elevations, making them inadequate in dynamic environments. Clearsky-Free approaches overcome these challenges. This work  critiques in fact the flaws of the \cyr{\texttt{CSI}}, particularly its multiplicative framework, which leads to division by zero and timestamp inconsistencies near sunrise and sunset. Note that the filtering data process using a solar angle threshold of 10–20° is insufficiently rigorous. Clearsky-Free methods for sure, or maybe additive models, offer simpler and more robust alternatives. Future improvements could include integrating high-quality  Clearsky models (like \cyr{\texttt{McClear}} or \cyr{\texttt{CAMS}}) directly as input of machine learning frameworks for enhanced accuracy and resilience, paving the way for more effective solar irradiance forecasting systems. The stationarization problem is significantly more complex for the tilted component of global irradiance (GTI), as it involves not only the Clearsky model error but also the additional error introduced by the transposition model. Clearsky-free models can be highly promising for predicting this component, which is directly linked to Photovoltaic production. This will be the focus of future works to explore this possibility further. 
\cyr{Collaboration between quality control and advanced Clearsky-Free methodologies significantly improves solar irradiance prediction under challenging atmospheric conditions or sparse data scenarios. While the proposed approach remains statistically viable for operational use, particularly under typical atmospheric conditions, extreme pollution events may challenge its accuracy. In such cases, integrating \cyr{\texttt{AOD}}-sensitive models or hybrid approaches could enhance robustness and adaptability. Striking a balance between simplicity, predictive power, and computational efficiency remains a key objective for future developments in Clearsky-Free forecasting frameworks.}
The implications of these developments are vast. Integrating these methods into decision-support systems for smart grid management, energy trading, and renewable energy integration offers transformative potential. The clears-sky free approach also could support more accurate direct and diffuse solar radiation forecasts, notably for inclined surfaces where applying clear-sky models is complicated. Real-time, accurate, and reliable forecasts can optimize solar energy usage in modern energy systems, contributing to efficient grid operations and accelerating the transition toward a low-carbon energy future. 

\section*{CRediT Authorship Contribution Statement}
CV: Writing – original draft, Methodology, Investigation, Formal analysis, Conceptualization. MD: Writing – review \& editing, Methodology. GN: Writing – review \& editing, Supervision. YMSD: review \& editing, Methodology. MA: Validation, Software, Data curation. LGG: Investigation, Data curation.

\appendix
\section{About Stationarity in Solar Energy Forecasting Context}
\label{Statio}
In time series analysis, stationarity is a critical concept referring to a process whose statistical properties, such as mean $\mathbb{E}[y_t]$, variance $\text{Var}(y_t)$, and autocovariance $\text{Cov}(y_t, y_{t+h})$, remain invariant over time \citep{voyant2017machine}. A stationary series satisfies:
\begin{equation}
\mathbb{E}[y_t] = \mu, \quad \text{Var}(y_t) = \sigma^2, \quad \text{and} \quad \text{Cov}(y_t, y_{t+h}) = \gamma(h),
\end{equation}
where $\mu$ and $\sigma^2$ are constants, and $\gamma(h)$ depends only on the lag $h$. Stationarity simplifies statistical modeling by ensuring that the estimated relationships within the data are consistent over time, a prerequisite for many traditional models like \cyr{\texttt{ARIMA}}.
In the context of solar energy forecasting, time series data are often non-stationary due to periodic components (daily and seasonal cycles), trends, and noise arising from atmospheric conditions. Transformations such as differencing ($y_t - y_{t-1}$) or detrending ($y_t - T(t)$ or $y_t / T(t)$ where $T(t)$ models a deterministic trend) are conventionally employed to enforce stationarity.
However, these transformations risk discarding valuable information intrinsic to the series, such as inherent seasonal patterns or long-term dependencies, particularly in univariate settings.
Recent advancements in machine learning \citep{Liu2023}, such as Extreme Learning Machines (\cyr{\texttt{ELM}}), challenge the necessity of strict stationarity, leveragging the raw structure of the data without requiring explicit stationarization \citep{Rahman2021, Silva2021, Liu2022}. The \cyr{\texttt{ELM}} training process, relying on a pseudo-inverse solution for $\boldsymbol{\beta}$ (see Section \ref{sec:ModelDescription}), is robust to non-stationary inputs, capturing periodic and stochastic variations directly within the endogenous structure of the series \citep{Pan2013}.
In univariate solar irradiance forecasting, the seasonal and diurnal cycles are embedded within the series itself. By feeding p-lagged observations $\{y_{t-1}, y_{t-2}, \dots, y_{t-p}\}$ as input to \cyr{\texttt{ELM}}s, these models can inherently learn the temporal dependencies and periodicities without explicit detrending or differencing. This approach aligns with the view that stationarization may not be essential in modern forecasting paradigms, particularly when the model architecture itself is designed to extract complex patterns from raw data.
Studies show that \cyr{\texttt{ELM}} deliver good accuracy while requiring significantly fewer resources compared to other deep learning models. This makes them particularly effective in operational short-term solar forecasting tasks, achieving good results with greatly reduced computational overhead \citep{en11102725}. The ability to handle non-stationary data directly underscores the potential of \cyr{\texttt{ELM}}s to maybe, redefine conventional practices in renewable energy forecasting.

\cyr{\section{Overview of the Extreme Learning Machine (\cyr{\texttt{ELM}}) for Solar Forecasting}
\label{annex:elm}
The \cyr{\texttt{ELM}} architecture consists of three layers: an input layer with $N_{\text{input}}$ neurons, a hidden layer with $N_{\text{hidden}}$ neurons, and a single-neuron output layer. For the 30-minute forecasting horizon, we use $N_{\text{input}} = 78$ historical features and $N_{\text{hidden}} = 472$ neurons in the hidden layer.
The model operates as follows:
\begin{itemize}
    \item Weight Initialization: Input-to-hidden weights $\mathbf{W} \in \mathbb{R}^{N_{\text{hidden}} \times N_{\text{input}}}$ are randomly drawn from a uniform distribution
    $W_{ij} \sim \mathcal{U}(-1,1).$
    A bias term $\mathbf{b} \in \mathbb{R}^{N_{\text{hidden}}}$ is also initialized randomly;
    \item Hidden Layer Activation: Each hidden neuron applies a nonlinear transformation to the input, with a mix of 60\% sigmoidal activation and 40\% Gaussian radial basis activation 
$    H_i = g(W_i X + b_i),$
    where $g(\cdot)$ is Sigmoid Activation (60\%; $ \frac{1}{1 + e^{-x}}$) or Gaussian Activation (40\%; $\exp(-\frac{x^2}{2\sigma^2}) \, \text{where} \, \sigma$ is set to 1 in our implementation.
    The proportion of sigmoid and Gaussian neurons is controlled by the threshold parameter $T = 0.6$;
    \item Output Weight Computation: Unlike conventional neural networks, \cyr{\texttt{ELM}} does not require backpropagation. Instead, output weights $\mathbf{\beta}$ are computed analytically using Ridge Regression:
    \begin{equation}
    \mathbf{\beta} = (H^T H + \lambda I)^{-1} H^T Y,
    \end{equation}
    where $H$ is the hidden layer matrix and $\lambda = 0.2$ is a regularization parameter;
    \item Training Procedure: To improve robustness and mitigate variance due to weight initialization, we conduct 200 independent runs ($j\in[1,200]$) with different initializations of $\mathbf{W}$ and $\mathbf{b}$. The best model is selected using a winner-takes-all strategy based on in-sample validation performance:
    \begin{equation}
    \mathbf{\beta}^* = \underset{\mathbf{\beta}_j}{\operatorname{argmin}} \, \left(\text{nRMSE} (\mathbf{\beta}_j)\right);
    \end{equation}
    \item Final Model for Out-of-Sample Testing: The selected model is then evaluated on the out-of-sample test set and compared with benchmark models.
\end{itemize}
To further clarify the structure of \cyr{\texttt{ELM}}, an example of architecture is shown in Figure \ref{fig:elm}:
\begin{figure}[h]
    \centering
    \includegraphics[width=1\textwidth]{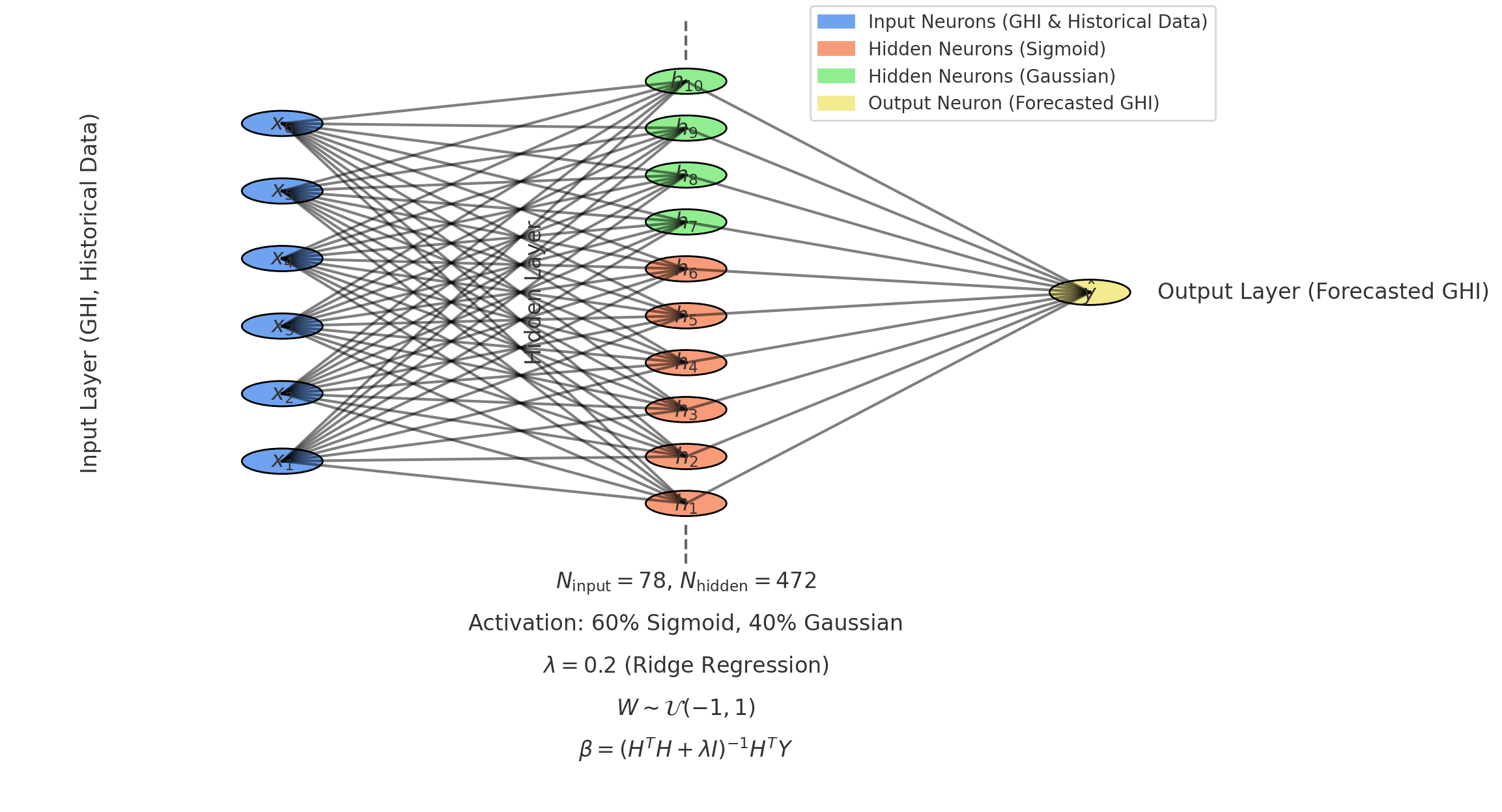}
    \caption{Schematic representation of the Extreme Learning Machine (\cyr{\texttt{ELM}}) architecture for a 30-minute forecast horizon. The model consists of an input layer ($N_{\text{input}} = 78$), a hidden layer ($N_{\text{hidden}} = 472$) with 60\% sigmoid and 40\% Gaussian neurons, and a single-output neuron predicting future \cyr{\texttt{GHI}}.}
    \label{fig:elm}
\end{figure}
The model consists of $ N_{\text{input}} = 78 $ input features, $ N_{\text{hidden}} = 472 $ hidden neurons, and a single output neuron, leading to a total of:
\begin{align}
(N_{\text{input}} + 1) \times N_{\text{hidden}} + (N_{\text{hidden}} + 1)& \times N_{\text{output}} = (78+1) \times 472 \notag \\
&\quad + (472+1) \times 1 = 37,401 \text{ parameters}.
\end{align}
Among these parameters, 37,368 parameters (99.91\%) are randomly initialized and remain fixed throughout training, and 473 parameters (1.26\%) are optimized using Ridge Regression.
The model includes four key hyperparameters: the number of hidden neurons $ N_{\text{hidden}} $, the activation mix ratio $ T $, the ridge regression regularization term $ \lambda $, and the number of runs $ N_{\text{runs}} $ for the winner-takes-all selection. 
Despite its analytical training process, \cyr{\texttt{ELM}} can be computationally intensive due to the matrix inversion in the output weight calculation, especially when scaling to larger datasets. Additionally, operational deployment may present challenges for energy operators, particularly in ensuring the robustness of predictions across diverse weather conditions, data availability, and the interpretability of the learned model. These aspects highlight the balance between model complexity and practical usability in experimental forecasting applications. Moreover, while our approach statistically learns the effects of atmospheric variations from historical data, it does not explicitly differentiate aerosol contributions nor adapt to sudden pollution peaks. This is a known limitation, as real-time AOD retrievals are not incorporated. However, given the rarity of extreme pollution events (fewer than five pollution alert days in 2024 in Nice City), this trade-off remains acceptable in the context of operational solar forecasting. Nonetheless, if such events were to become more frequent, future work could explore hybrid approaches integrating real-time aerosol information.}

\section{Non-Parametric Prediction Interval Generation}
\label{lookup}
In addition to Quantile Regression, a non-parametric methodology was implemented to derive prediction intervals directly from the residuals of deterministic forecasts \citep{Pinson2007, Wang2022}. This approach uses a lookup table generated during the training phase (in-sample data), which captures empirical relationships between prediction errors and desired coverage levels (between 0 and 1).
During the training phase, the residuals of deterministic forecasts ($ y - \hat{y}$) are computed. For each residual, the standard deviation $\hat{\sigma}$ is calculated and depend on the site and the horizon considered. For a given coverage level $(1-\alpha)$, prediction intervals are defined for out-sample data as $\underline{y}(\alpha) = \hat{y} - k_\alpha \times \hat{\sigma}, \quad \overline{y}(\alpha) = \hat{y} + k_\alpha \times \hat{\sigma}$, 
where $\hat{y}$ is the deterministic forecast, and $k_\alpha$ is a unitless scaling factor derived from the residual distribution to achieve the desired coverage. \cyr{\texttt{PICP}} and \cyr{\texttt{nMIL }}are then computed according equations in Section {\ref{metrics}}.
By interpolating these metrics across coverage levels $\alpha$, a lookup table is constructed (several values $k_\alpha^{lookup}$). This table maps each $\alpha$ to a corresponding scaling factor $k_\alpha$, enabling the generation of prediction intervals without assuming a specific error distribution.
During testing, the lookup table is used to compute prediction intervals for a given $\alpha$. For each forecast $\hat{y}$, the prediction intervals are derived as $\underline{y}(\alpha) = \hat{y} - k_\alpha^{lookup} \times \hat{\sigma}, \quad \overline{y}(\alpha) = \hat{y} +  k_\alpha^{lookup} \times \hat{\sigma}$.
This process yields prediction interval over a nominal coverage ($1-\alpha$) according $[\underline{y}(\alpha)< \hat{y}< \overline{y}(\alpha)]_{(1-\alpha)}$.
The lookup-table-based methodology provides several notable advantages. By avoiding assumptions about error distributions, it adapts flexibly to diverse atmospheric conditions, ensuring accurate predictions even in non-stationary environments. Its robustness stems from empirical relationships established between residuals and forecasts during the training phase, allowing for the dynamic generation of reliable prediction intervals.
Additionally, this non-parametric approach complements \cyr{\texttt{QR}} by offering a computationally efficient alternative for deriving probabilistic forecasts \citep{Perna2024}. Together, the Quantile Regression (\cyr{\texttt{QR}}) and the lookup table integrated with the \cyr{\texttt{ELM}} model, offer a scalable and practical solution for operational solar irradiance prediction, capable of delivering probabilistic forecasts.
\begin{figure}
    \centering
    \includegraphics[width=0.5\textwidth]{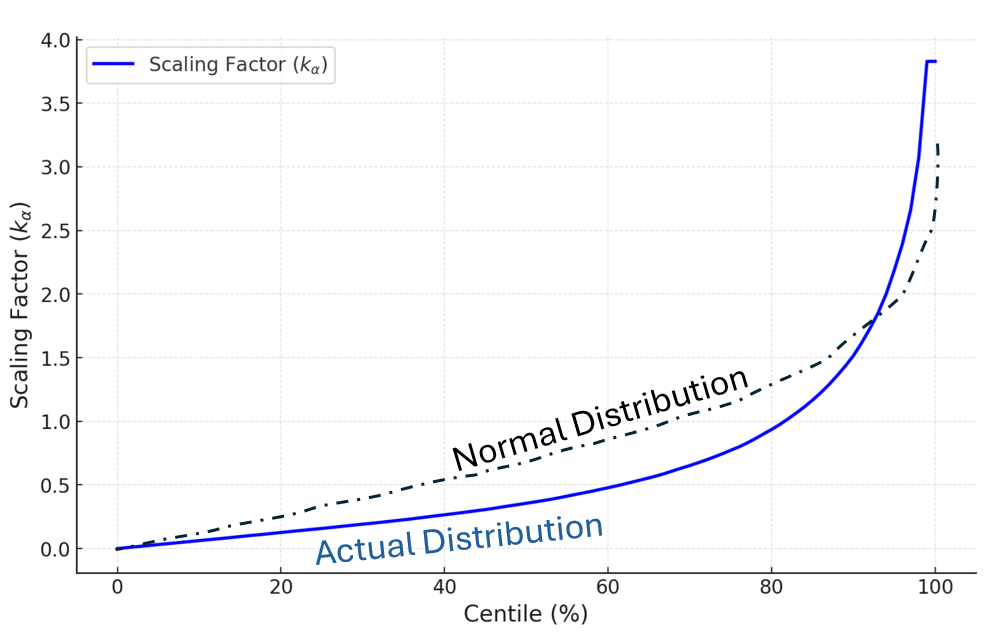}
    \caption{Lookup table representing scaling factors $k_\alpha$ (mean of all stations concerning 30 min horizon) across centiles, used to compute prediction intervals for probabilistic forecasts.}
    \label{fig:lookup_table}
\end{figure}
The lookup table in Figure~\ref{fig:lookup_table} illustrates the scaling factors ($k_\alpha$) derived during the training phase for each centile $(0\%$ to $100\%$). The curve is smooth, reflecting a well-defined relationship between residual variance and coverage levels. Lower centiles show minimal scaling, while higher centiles exhibit increasing values, consistent with broader prediction intervals required for higher coverage. This structure ensures a dynamic adjustment of prediction intervals, allowing for robust probabilistic forecasts under varying atmospheric conditions. It will be noted that $k_{\alpha} = 2$ is close to providing a nominal coverage of 95\%, as in the parametric Gaussian case.

\section{Quantile-Based Approximation of \cyr{\texttt{CRPS}}}
\label{CRPS}
In this study, \cyr{\texttt{CRPS}} is approximated using quantiles instead of the full \cyr{\texttt{CDF}} \citep{Broecker2012}. Let $Q_\alpha$ denote the predicted quantile at probability level $(1-\alpha)$, such that $Q_\alpha$ satisfies
$ P(Y \leq Q_\alpha) = \alpha,  \text{for } \alpha \in [0, 1].$
The quantile-based \cyr{\texttt{CRPS}} is computed as the weighted sum of absolute deviations between the predicted quantiles $Q_\alpha$ and the observation $y$ via 
\cyr{\texttt{CRPS}}$(Q, y) = \sum_{i=1}^{N_\alpha} w_i \cdot \big| Q_{\alpha_i} - y \big|,$
where $N_\alpha$ is the number of quantiles used (here, $N_\alpha = 101$), and the weights $w_i$ correspond to the quantile resolution ($w_i = \Delta \alpha = 0.01$ for evenly spaced quantiles). This reduces to \cyr{\texttt{CRPS}}$(Q, y) = 0.01 \cdot \sum_{i=1}^{101} \big| Q_{\alpha_i} - y \big|.$
The classical and quantile-based \cyr{\texttt{CRPS}} formulations are related through the fact that the \cyr{\texttt{CDF}} can be approximated by quantiles using $F(x) \approx \sum_{\alpha_i \leq x} w_i.$
Thus, the quantile-based \cyr{\texttt{CRPS}} is a discrete approximation of the continuous integral, where the quantiles $Q_\alpha$ replace the explicit \cyr{\texttt{CDF}}. The classical \cyr{\texttt{CRPS}} provides a precise evaluation when a continuous \cyr{\texttt{CDF}} is available, making it suitable for models that predict full distributions, such as Gaussian processes. In contrast, the quantile-based \cyr{\texttt{CRPS}} is particularly efficient for models like Quantile Regression, which inherently provide discrete quantile estimates. This approach avoids the need to reconstruct or approximate the \cyr{\texttt{CDF}}, offering a simpler yet robust alternative.
The quantile-based \cyr{\texttt{CRPS}} also provides practical advantages. By avoiding integration over a continuous distribution, it achieves computational efficiency, making it well-suited for large datasets. Furthermore, it aligns naturally with quantile-based models, such as Quantile Regression, without requiring additional assumptions. This method directly handles probabilistic forecasts expressed as quantiles, simplifying the evaluation pipeline.
The accuracy of the quantile-based approximation depends on the resolution of the quantiles \citep{Zamo2018}. For evenly spaced quantiles, as in this study ($101$ quantiles), the approximation is sufficiently precise for practical purposes. To ensure interpretability, a normalized \cyr{\texttt{CRPS}} is also computed as \cyr{\texttt{nCRPS}} $= CRPS \cdot {\text{E}[y]}^{-1}.$
This normalization enables a relative comparison across datasets with different scales.

\section{Analysis of Models Performance}
\label{bench}
\begin{figure}
    \centering
    \includegraphics[width=0.9\textwidth]{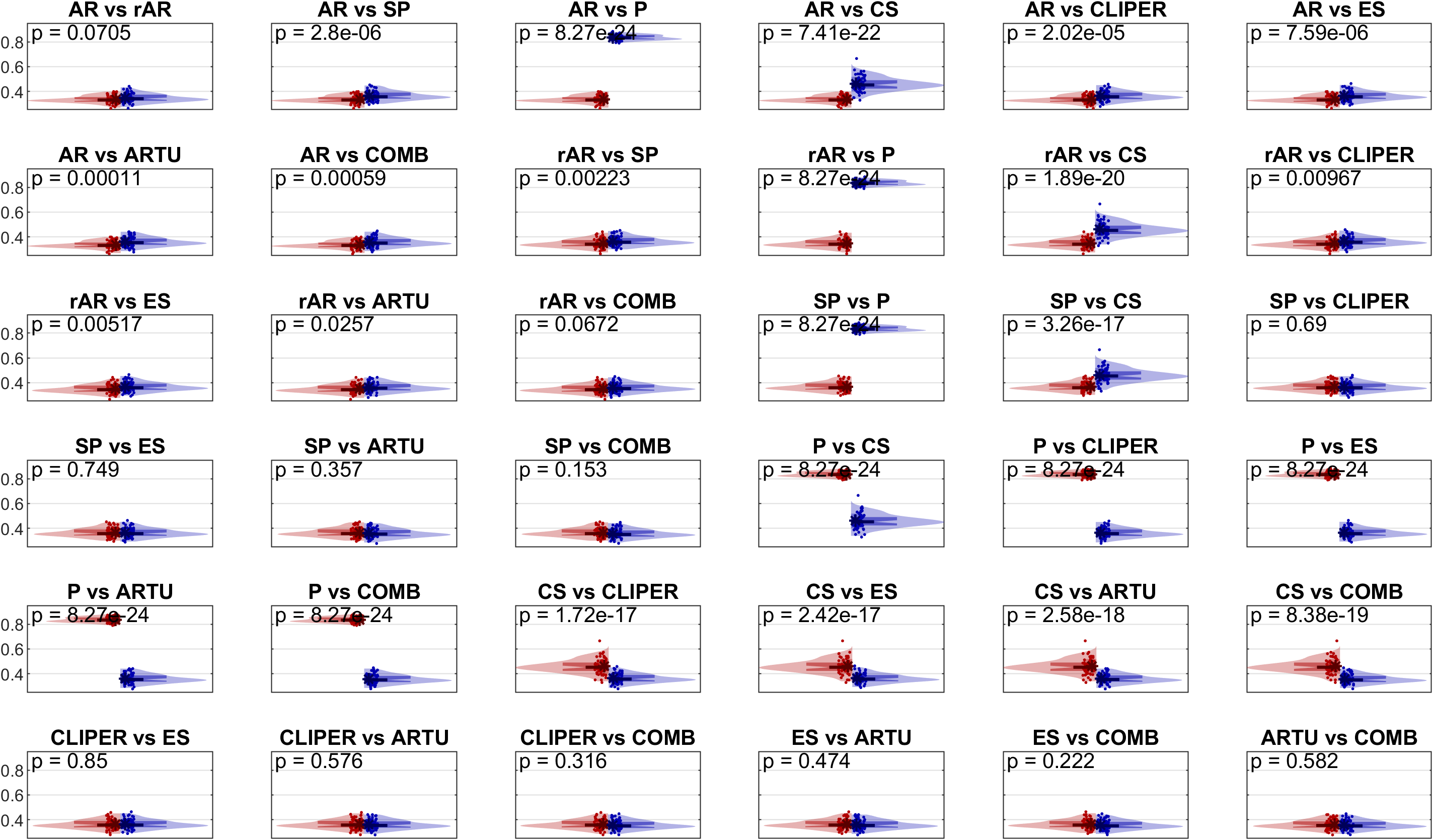}
    \caption{Cross-comparison of benchmark methods for all sites and a 180-min horizon. Each subplot shows \cyr{\texttt{nRMSE}} distributions for a pair of methods, with $p$-values indicating the significance of the difference.}
    \label{fig:c1_v1}
\end{figure}

Figure~\ref{fig:c1_v1} compares the normalized Root Mean Square Error (\cyr{\texttt{nRMSE}}) distributions of the evaluated methods. \cyr{\texttt{AR}} consistently outperforms most models, including \cyr{\texttt{COMB}} ($p = 0.00059$) and \cyr{\texttt{SP}} ($p = 2.8 \times 10^{-6}$), with distributions centered at lower \cyr{\texttt{nRMSE}} values. While \cyr{\texttt{AR}} and \cyr{\texttt{rAR}} show near-equivalent performance ($p = 0.0705$), \cyr{\texttt{AR}} holds a slight edge with lower median errors.
COMB and \cyr{\texttt{SP}} are competitive but fail to match \cyr{\texttt{AR}}, especially against weaker methods like \cyr{\texttt{P}} and \cyr{\texttt{CS}}, which consistently rank as the poorest performers ($p = 8.27 \times 10^{-24}$ and $7.41 \times 10^{-22}$, respectively). \cyr{\texttt{CLIPER}} and \cyr{\texttt{ES}} demonstrate intermediate performance, with no significant differences between them ($p = 0.85$).
Overall, \cyr{\texttt{AR}} emerges as the most reliable method, closely followed by \cyr{\texttt{rAR}}. \cyr{\texttt{CLIPER}} and \cyr{\texttt{SP}} provide simpler, yet competitive alternatives, while \cyr{\texttt{COMB}}, \cyr{\texttt{ES}}, and \cyr{\texttt{ARTU}} offer strong but more sophisticated solutions. \cyr{\texttt{P}} and \cyr{\texttt{CS}} remain the least effective models.

\bibliography{Main}

\begin{thebibliography}{51}
\expandafter\ifx\csname natexlab\endcsname\relax\def\natexlab#1{#1}\fi
\providecommand{\url}[1]{\texttt{#1}}
\providecommand{\href}[2]{#2}
\providecommand{\path}[1]{#1}
\providecommand{\DOIprefix}{doi:}
\providecommand{\ArXivprefix}{arXiv:}
\providecommand{\URLprefix}{URL: }
\providecommand{\Pubmedprefix}{pmid:}
\providecommand{\doi}[1]{\href{http://dx.doi.org/#1}{\path{#1}}}
\providecommand{\Pubmed}[1]{\href{pmid:#1}{\path{#1}}}
\providecommand{\bibinfo}[2]{#2}
\ifx\xfnm\relax \def\xfnm[#1]{\unskip,\space#1}\fi
\bibitem[{Akarslan et~al.(2018)Akarslan, Hocaoglu and
  Edizkan}]{akarslan2018novel}
\bibinfo{author}{Akarslan, E.}, \bibinfo{author}{Hocaoglu, F.},
  \bibinfo{author}{Edizkan, R.}, \bibinfo{year}{2018}.
\newblock \bibinfo{title}{Novel short term solar irradiance forecasting
  models}.
\newblock \bibinfo{journal}{Renew Energy} \bibinfo{volume}{123},
  \bibinfo{pages}{58--66}.
\newblock \DOIprefix\doi{10.1016/J.RENENE.2018.02.048}.
\bibitem[{Al-Dahidi et~al.(2018)Al-Dahidi, Ayadi, Adeeb, Alrbai and
  Qawasmeh}]{en11102725}
\bibinfo{author}{Al-Dahidi, S.}, \bibinfo{author}{Ayadi, O.},
  \bibinfo{author}{Adeeb, J.}, \bibinfo{author}{Alrbai, M.},
  \bibinfo{author}{Qawasmeh, B.R.}, \bibinfo{year}{2018}.
\newblock \bibinfo{title}{Extreme learning machines for solar photovoltaic
  power predictions}.
\newblock \bibinfo{journal}{Energies} \bibinfo{volume}{11}.
\newblock \URLprefix \url{https://www.mdpi.com/1996-1073/11/10/2725},
  \DOIprefix\doi{10.3390/en11102725}.
\bibitem[{Allen et~al.(1998)Allen, Pereira, Raes and Smith}]{allen1998crop}
\bibinfo{author}{Allen, R.G.}, \bibinfo{author}{Pereira, L.S.},
  \bibinfo{author}{Raes, D.}, \bibinfo{author}{Smith, M.},
  \bibinfo{year}{1998}.
\newblock \bibinfo{title}{Crop evapotranspiration: Guidelines for computing
  crop water requirements}.
\newblock \bibinfo{journal}{FAO Irrigation and Drainage Paper}
  \bibinfo{volume}{56}, \bibinfo{pages}{1--300}.
\newblock \URLprefix \url{www.fao.org/4/x0490e/x0490e00.htm}.
\bibitem[{Azarpour et~al.(2022)Azarpour, Mohammadzadeh, Rezaei and
  Zendehboudi}]{ren2015review}
\bibinfo{author}{Azarpour, A.}, \bibinfo{author}{Mohammadzadeh, O.},
  \bibinfo{author}{Rezaei, N.}, \bibinfo{author}{Zendehboudi, S.},
  \bibinfo{year}{2022}.
\newblock \bibinfo{title}{Current status and future prospects of renewable and
  sustainable energy in north america: Progress and challenges}.
\newblock \bibinfo{journal}{Energy Conversion and Management}
  \bibinfo{volume}{269}, \bibinfo{pages}{115945}.
\newblock \DOIprefix\doi{10.1016/j.enconman.2022.115945}.
\bibitem[{Bröcker(2012)}]{Broecker2012}
\bibinfo{author}{Bröcker, J.}, \bibinfo{year}{2012}.
\newblock \bibinfo{title}{Evaluating raw ensembles with the continuous ranked
  probability score}.
\newblock \bibinfo{journal}{Quarterly Journal of the Royal Meteorological
  Society} \bibinfo{volume}{138}.
\newblock \DOIprefix\doi{10.1002/qj.1891}.
\bibitem[{Chodakowska et~al.(2024)Chodakowska, Nazarko, Nazarko and
  Rabayah}]{wen2022data}
\bibinfo{author}{Chodakowska, E.}, \bibinfo{author}{Nazarko, J.},
  \bibinfo{author}{Nazarko, {\L}.}, \bibinfo{author}{Rabayah, H.S.},
  \bibinfo{year}{2024}.
\newblock \bibinfo{title}{Solar radiation forecasting: A systematic meta-review
  of current methods and emerging trends}.
\newblock \bibinfo{journal}{Energies} \bibinfo{volume}{17}.
\newblock \DOIprefix\doi{10.3390/en17133156}.
\bibitem[{David et~al.(2016)David, Ramahatana, Trombe and
  Lauret}]{david2016probabilistic}
\bibinfo{author}{David, M.}, \bibinfo{author}{Ramahatana, F.},
  \bibinfo{author}{Trombe, P.}, \bibinfo{author}{Lauret, P.},
  \bibinfo{year}{2016}.
\newblock \bibinfo{title}{Probabilistic forecasting of the solar irradiance
  with recursive arma and garch models}.
\newblock \bibinfo{journal}{Solar Energy} \bibinfo{volume}{133},
  \bibinfo{pages}{55--72}.
\newblock \DOIprefix\doi{10.1016/J.SOLENER.2016.03.064}.
\bibitem[{Despotovic et~al.(2024)Despotovic, Voyant, Garcia-Gutierrez, Almorox
  and Notton}]{DESPOTOVIC2024123215}
\bibinfo{author}{Despotovic, M.}, \bibinfo{author}{Voyant, C.},
  \bibinfo{author}{Garcia-Gutierrez, L.}, \bibinfo{author}{Almorox, J.},
  \bibinfo{author}{Notton, G.}, \bibinfo{year}{2024}.
\newblock \bibinfo{title}{Solar irradiance time series forecasting using
  auto-regressive and extreme learning methods: Influence of transfer learning
  and clustering}.
\newblock \bibinfo{journal}{Applied Energy} \bibinfo{volume}{365},
  \bibinfo{pages}{123215}.
\newblock \DOIprefix\doi{10.1016/j.apenergy.2024.123215}.
\bibitem[{Diagne et~al.(2013)Diagne, David, Lauret, Boland and
  Schmutz}]{diagne2013review}
\bibinfo{author}{Diagne, M.}, \bibinfo{author}{David, M.},
  \bibinfo{author}{Lauret, P.}, \bibinfo{author}{Boland, J.},
  \bibinfo{author}{Schmutz, N.}, \bibinfo{year}{2013}.
\newblock \bibinfo{title}{Review of solar irradiance forecasting methods and a
  proposition for small-scale insular grids}.
\newblock \bibinfo{journal}{Renewable and Sustainable Energy Reviews}
  \bibinfo{volume}{27}, \bibinfo{pages}{65--76}.
\newblock \DOIprefix\doi{10.1016/j.rser.2013.06.042}.
\bibitem[{Frimane et~al.(2022)Frimane, Munkhammar and van~der
  Meer}]{frimane2022infinite}
\bibinfo{author}{Frimane, {\^A}.}, \bibinfo{author}{Munkhammar, J.},
  \bibinfo{author}{van~der Meer, D.}, \bibinfo{year}{2022}.
\newblock \bibinfo{title}{Infinite hidden markov model for short-term solar
  irradiance forecasting}.
\newblock \bibinfo{journal}{Solar Energy} \bibinfo{volume}{244},
  \bibinfo{pages}{331--342}.
\newblock \DOIprefix\doi{10.1016/J.SOLENER.2022.08.041}.
\bibitem[{Garcia-Gutierrez et~al.(2022)Garcia-Gutierrez, Voyant, Notton and
  Almorox}]{app12178529}
\bibinfo{author}{Garcia-Gutierrez, L.}, \bibinfo{author}{Voyant, C.},
  \bibinfo{author}{Notton, G.}, \bibinfo{author}{Almorox, J.},
  \bibinfo{year}{2022}.
\newblock \bibinfo{title}{Evaluation and comparison of spatial clustering for
  solar irradiance time series}.
\newblock \bibinfo{journal}{Applied Sciences} \bibinfo{volume}{12}.
\newblock \DOIprefix\doi{10.3390/app12178529}.
\bibitem[{Gneiting et~al.(2023)Gneiting, Lerch and
  Schulz}]{gneiting2023probabilistic}
\bibinfo{author}{Gneiting, T.}, \bibinfo{author}{Lerch, S.},
  \bibinfo{author}{Schulz, B.}, \bibinfo{year}{2023}.
\newblock \bibinfo{title}{Probabilistic solar forecasting: Benchmarks,
  post-processing, verification}.
\newblock \bibinfo{journal}{Solar Energy} \bibinfo{volume}{252},
  \bibinfo{pages}{72--80}.
\newblock \DOIprefix\doi{10.1016/J.SOLENER.2022.12.054}.
\bibitem[{Gneiting and Raftery(2007)}]{gneiting2007strictly}
\bibinfo{author}{Gneiting, T.}, \bibinfo{author}{Raftery, A.E.},
  \bibinfo{year}{2007}.
\newblock \bibinfo{title}{Strictly proper scoring rules, prediction, and
  estimation}.
\newblock \bibinfo{journal}{Journal of the American Statistical Association}
  \bibinfo{volume}{102}, \bibinfo{pages}{359--378}.
\newblock \DOIprefix\doi{10.1198/016214506000001437}.
\bibitem[{Gueymard(2008)}]{gueymard2021solar}
\bibinfo{author}{Gueymard, C.A.}, \bibinfo{year}{2008}.
\newblock \bibinfo{title}{Rest2: High-performance solar radiation model for
  cloudless-sky irradiance, illuminance, and photosynthetically active
  radiation – validation with a benchmark dataset}.
\newblock \bibinfo{journal}{Solar Energy} \bibinfo{volume}{82},
  \bibinfo{pages}{272--285}.
\newblock \DOIprefix\doi{10.1016/j.solener.2007.04.008}.
\bibitem[{Haixu et~al.(2022)Haixu, Wu, Jianmin and Mingsheng}]{Liu2022}
\bibinfo{author}{Haixu, Y.L.}, \bibinfo{author}{Wu}, \bibinfo{author}{Jianmin,
  W.}, \bibinfo{author}{Mingsheng, L.}, \bibinfo{year}{2022}.
\newblock \bibinfo{title}{Non-stationary transformers: Rethinking the
  stationarity in time series forecasting}.
\newblock \bibinfo{journal}{Advances in Neural Information Processing Systems}
  \bibinfo{volume}{35}.
\newblock \DOIprefix\doi{10.48550/arXiv.2205.14415}.
\bibitem[{Herrería-Alonso et~al.(2020)Herrería-Alonso, Suárez-González,
  Rodríguez-Pérez, Rodríguez-Rubio and
  López-García}]{cervantes2019evaluation}
\bibinfo{author}{Herrería-Alonso, S.}, \bibinfo{author}{Suárez-González,
  A.}, \bibinfo{author}{Rodríguez-Pérez, M.},
  \bibinfo{author}{Rodríguez-Rubio, R.F.}, \bibinfo{author}{López-García,
  C.}, \bibinfo{year}{2020}.
\newblock \bibinfo{title}{A solar altitude angle model for efficient solar
  energy predictions}.
\newblock \bibinfo{journal}{Sensors} \bibinfo{volume}{20}.
\newblock \DOIprefix\doi{10.3390/s20051391}.
\bibitem[{Hoyos-Gómez et~al.(2022)Hoyos-Gómez, Ruiz-Muñoz and
  Ruiz-Mendoza}]{hoyos2022short}
\bibinfo{author}{Hoyos-Gómez, L.}, \bibinfo{author}{Ruiz-Muñoz, J.},
  \bibinfo{author}{Ruiz-Mendoza, B.}, \bibinfo{year}{2022}.
\newblock \bibinfo{title}{Short-term forecasting of global solar irradiance in
  tropical environments with incomplete data}.
\newblock \bibinfo{journal}{Appl Energy} \bibinfo{volume}{307},
  \bibinfo{pages}{118192}.
\newblock \DOIprefix\doi{10.1016/J.APENERGY.2021.118192}.
\bibitem[{Huang et~al.(2006)Huang, Zhu and Siew}]{huang2006extreme}
\bibinfo{author}{Huang, G.B.}, \bibinfo{author}{Zhu, Q.Y.},
  \bibinfo{author}{Siew, C.K.}, \bibinfo{year}{2006}.
\newblock \bibinfo{title}{Extreme learning machine: Theory and applications}.
\newblock \bibinfo{journal}{Neurocomputing} \bibinfo{volume}{70},
  \bibinfo{pages}{489--501}.
\newblock \DOIprefix\doi{10.1016/j.neucom.2005.12.126}.
\bibitem[{{International Energy Agency (IEA)}(2021)}]{iea2021solar}
\bibinfo{author}{{International Energy Agency (IEA)}}, \bibinfo{year}{2021}.
\newblock \bibinfo{title}{Renewables 2021: Analysis and forecasts to 2026}.
\newblock \bibinfo{journal}{IEA, Paris} \URLprefix
  \url{www.iea.org/reports/renewables-2021}. \bibinfo{note}{licence: CC BY
  4.0}.
\bibitem[{Koenker(2005)}]{koenker2005quantile}
\bibinfo{author}{Koenker, R.}, \bibinfo{year}{2005}.
\newblock \bibinfo{title}{Quantile Regression}.
\newblock \bibinfo{publisher}{Cambridge University Press}.
\newblock \DOIprefix\doi{10.1017/CBO9780511754098}.
\bibitem[{Kumar et~al.(2020)Kumar, Yagli, Kashyap and
  Srinivasan}]{jiang2021comparison}
\bibinfo{author}{Kumar, D.S.}, \bibinfo{author}{Yagli, G.M.},
  \bibinfo{author}{Kashyap, M.}, \bibinfo{author}{Srinivasan, D.},
  \bibinfo{year}{2020}.
\newblock \bibinfo{title}{Solar irradiance resource and forecasting: a
  comprehensive review}.
\newblock \bibinfo{journal}{IET Renewable Power Generation}
  \bibinfo{volume}{14}, \bibinfo{pages}{1641--1656}.
\newblock \DOIprefix\doi{10.1049/iet-rpg.2019.1227}.
\bibitem[{Lauret et~al.(2022)Lauret, Alonso-Suárez, Le~Gal, Salle and
  David}]{lauret2022forecasts}
\bibinfo{author}{Lauret, P.}, \bibinfo{author}{Alonso-Suárez, R.},
  \bibinfo{author}{Le~Gal, J.}, \bibinfo{author}{Salle, L.},
  \bibinfo{author}{David, M.}, \bibinfo{year}{2022}.
\newblock \bibinfo{title}{Solar forecasts based on the clear sky index or the
  clearness index: Which is better?}
\newblock \bibinfo{journal}{Solar} \bibinfo{volume}{2},
  \bibinfo{pages}{432--444}.
\newblock \DOIprefix\doi{10.3390/SOLAR2040026}.
\bibitem[{Lauret et~al.(2017)Lauret, David and Pedro}]{martin2021probabilistic}
\bibinfo{author}{Lauret, P.}, \bibinfo{author}{David, M.},
  \bibinfo{author}{Pedro, H.T.C.}, \bibinfo{year}{2017}.
\newblock \bibinfo{title}{Probabilistic solar forecasting using quantile
  regression models}.
\newblock \bibinfo{journal}{Energies} \bibinfo{volume}{10}.
\newblock \DOIprefix\doi{10.3390/en10101591}.
\bibitem[{Lauret et~al.(2012)Lauret, Rodler, Muselli, David, Diagne and
  Voyant}]{lauret2012bayesianmodelcommitteeapproach}
\bibinfo{author}{Lauret, P.}, \bibinfo{author}{Rodler, A.},
  \bibinfo{author}{Muselli, M.}, \bibinfo{author}{David, M.},
  \bibinfo{author}{Diagne, H.M.}, \bibinfo{author}{Voyant, C.},
  \bibinfo{year}{2012}.
\newblock \bibinfo{title}{A b\lowercase{AYESIAN MODEL COMMITTEE APPROACH TO
  FORECASTING GLOBAL SOLAR RADIATION}}, in: \bibinfo{booktitle}{{WREF 2012 :
  World Renewable Energy Forum}}, \bibinfo{address}{Denver, United States}.
  p.~\bibinfo{pages}{1}.
\newblock \URLprefix \url{https://hal.science/hal-00682217}.
\bibitem[{{Lef{\`e}vre} et~al.(2013){Lef{\`e}vre}, {Oumbe}, {Blanc}, {Espinar},
  {Gschwind}, {Qu}, {Wald}, {Schroedter-Homscheidt}, {Hoyer-Klick}, {Arola},
  {Benedetti}, {Kaiser} and {Morcrette}}]{lefevre2013mcclear}
\bibinfo{author}{{Lef{\`e}vre}, M.}, \bibinfo{author}{{Oumbe}, A.},
  \bibinfo{author}{{Blanc}, P.}, \bibinfo{author}{{Espinar}, B.},
  \bibinfo{author}{{Gschwind}, B.}, \bibinfo{author}{{Qu}, Z.},
  \bibinfo{author}{{Wald}, L.}, \bibinfo{author}{{Schroedter-Homscheidt}, M.},
  \bibinfo{author}{{Hoyer-Klick}, C.}, \bibinfo{author}{{Arola}, A.},
  \bibinfo{author}{{Benedetti}, A.}, \bibinfo{author}{{Kaiser}, J.W.},
  \bibinfo{author}{{Morcrette}, J.J.}, \bibinfo{year}{2013}.
\newblock \bibinfo{title}{{McClear: a new model estimating downwelling solar
  radiation at ground level in clear-sky conditions}}.
\newblock \bibinfo{journal}{Atmospheric Measurement Techniques}
  \bibinfo{volume}{6}, \bibinfo{pages}{2403--2418}.
\newblock \DOIprefix\doi{10.5194/amt-6-2403-2013}.
\bibitem[{Li et~al.(2016)Li, Lou, Lam and Wu}]{Danny2016}
\bibinfo{author}{Li, D.H.}, \bibinfo{author}{Lou, S.}, \bibinfo{author}{Lam,
  J.C.}, \bibinfo{author}{Wu, R.H.}, \bibinfo{year}{2016}.
\newblock \bibinfo{title}{Determining solar irradiance on inclined planes from
  classified {CIE} (international commission on illumination) standard skies}.
\newblock \bibinfo{journal}{Energy} \bibinfo{volume}{101},
  \bibinfo{pages}{462--470}.
\newblock \DOIprefix\doi{10.1016/j.energy.2016.02.054}.
\bibitem[{Liu et~al.(2024)Liu, Cheng, Li, Huang, Liu, Xie and Chen}]{Liu2023}
\bibinfo{author}{Liu, Z.}, \bibinfo{author}{Cheng, M.}, \bibinfo{author}{Li,
  Z.}, \bibinfo{author}{Huang, Z.}, \bibinfo{author}{Liu, Q.},
  \bibinfo{author}{Xie, Y.}, \bibinfo{author}{Chen, E.}, \bibinfo{year}{2024}.
\newblock \bibinfo{title}{Adaptive normalization for non-stationary time series
  forecasting: A temporal slice perspective}.
\newblock \bibinfo{journal}{Advances in Neural Information Processing Systems}
  \bibinfo{volume}{36}.
\newblock \DOIprefix\doi{10.5555/3666122.3666750}.
\bibitem[{Ma et~al.(2020)Ma, Zhang, Mei, Zhen, Gao and Zhou}]{blanc2019short}
\bibinfo{author}{Ma, Y.}, \bibinfo{author}{Zhang, X.}, \bibinfo{author}{Mei,
  S.}, \bibinfo{author}{Zhen, Z.}, \bibinfo{author}{Gao, R.},
  \bibinfo{author}{Zhou, Z.}, \bibinfo{year}{2020}.
\newblock \bibinfo{title}{Ultra-short-term solar power forecasting based on a
  modified clear sky model}, in: \bibinfo{booktitle}{2020 39th Chinese Control
  Conference (CCC)}, pp. \bibinfo{pages}{5311--5316}.
\newblock \DOIprefix\doi{10.23919/CCC50068.2020.9189533}.
\bibitem[{Makridakis et~al.(1998)Makridakis, Wheelwright and
  Hyndman}]{makridakis1998forecasting}
\bibinfo{author}{Makridakis, S.}, \bibinfo{author}{Wheelwright, S.C.},
  \bibinfo{author}{Hyndman, R.J.}, \bibinfo{year}{1998}.
\newblock \bibinfo{title}{Forecasting methods and applications}.
\newblock \bibinfo{edition}{3rd edition} ed., \bibinfo{publisher}{Wiley}.
\newblock \DOIprefix\doi{10.2307/2287014}.
\bibitem[{Mann and Whitney(1947)}]{Mann-Whitney-U-test}
\bibinfo{author}{Mann, H.B.}, \bibinfo{author}{Whitney, D.R.},
  \bibinfo{year}{1947}.
\newblock \bibinfo{title}{{On a Test of Whether one of Two Random Variables is
  Stochastically Larger than the Other}}.
\newblock \bibinfo{journal}{The Annals of Mathematical Statistics}
  \bibinfo{volume}{18}, \bibinfo{pages}{50 -- 60}.
\newblock \DOIprefix\doi{10.1214/aoms/1177730491}.
\bibitem[{Nelder and Mead(1965)}]{nelder1965simplex}
\bibinfo{author}{Nelder, J.A.}, \bibinfo{author}{Mead, R.},
  \bibinfo{year}{1965}.
\newblock \bibinfo{title}{A simplex method for function minimization}.
\newblock \bibinfo{journal}{The computer journal} \bibinfo{volume}{7},
  \bibinfo{pages}{308--313}.
\newblock \DOIprefix\doi{10.1093/comjnl/7.4.308}.
\bibitem[{Pan and Zhao(2013)}]{Pan2013}
\bibinfo{author}{Pan, F.}, \bibinfo{author}{Zhao, H.}, \bibinfo{year}{2013}.
\newblock \bibinfo{title}{Online sequential extreme learning machine based
  multilayer perception with output self feedback for time series prediction}.
\newblock \bibinfo{journal}{Journal of Shanghai Jiaotong University (Science)}
  \bibinfo{volume}{18}, \bibinfo{pages}{366--375}.
\newblock \DOIprefix\doi{10.1007/S12204-013-1407-0}.
\bibitem[{Perna et~al.(2024)Perna, Austnes, Gerini, Chevron, Fazio, Falco and
  Paolone}]{Perna2024}
\bibinfo{author}{Perna, S.}, \bibinfo{author}{Austnes, P.F.},
  \bibinfo{author}{Gerini, F.}, \bibinfo{author}{Chevron, M.},
  \bibinfo{author}{Fazio, A.D.}, \bibinfo{author}{Falco, P.D.},
  \bibinfo{author}{Paolone, M.}, \bibinfo{year}{2024}.
\newblock \bibinfo{title}{A comparative analysis of empirical copula and
  quantile regression methods for probabilistic load forecasting}, in:
  \bibinfo{booktitle}{2024 18th International Conference on Probabilistic
  Methods Applied to Power Systems (PMAPS)}, pp. \bibinfo{pages}{1--6}.
\newblock \DOIprefix\doi{10.1109/PMAPS61648.2024.10667333}.
\bibitem[{Pinson et~al.(2007)Pinson, Nielsen, Møller, Madsen and
  Kariniotakis}]{Pinson2007}
\bibinfo{author}{Pinson, P.}, \bibinfo{author}{Nielsen, H.},
  \bibinfo{author}{Møller, J.}, \bibinfo{author}{Madsen, H.},
  \bibinfo{author}{Kariniotakis, G.}, \bibinfo{year}{2007}.
\newblock \bibinfo{title}{Non‐parametric probabilistic forecasts of wind
  power: required properties and evaluation}.
\newblock \bibinfo{journal}{Wind Energy} \bibinfo{volume}{10},
  \bibinfo{pages}{497--516}.
\newblock \DOIprefix\doi{10.1002/WE.230}.
\bibitem[{Rahman et~al.(2021)Rahman, Shakeri, Tiong, Khatun, Amin, Pasupuleti
  and Hasan}]{Rahman2021}
\bibinfo{author}{Rahman, M.}, \bibinfo{author}{Shakeri, M.},
  \bibinfo{author}{Tiong, S.}, \bibinfo{author}{Khatun, F.},
  \bibinfo{author}{Amin, N.}, \bibinfo{author}{Pasupuleti, J.},
  \bibinfo{author}{Hasan, M.K.}, \bibinfo{year}{2021}.
\newblock \bibinfo{title}{Prospective methodologies in hybrid renewable energy
  systems for energy prediction using artificial neural networks}.
\newblock \bibinfo{journal}{Sustainability} \bibinfo{volume}{13},
  \bibinfo{pages}{2393}.
\newblock \DOIprefix\doi{10.3390/SU13042393}.
\bibitem[{Ruiz-Arias(2023)}]{ruiz-arias_sparta_2023}
\bibinfo{author}{Ruiz-Arias, J.A.}, \bibinfo{year}{2023}.
\newblock \bibinfo{title}{{SPARTA}: {Solar} parameterization for the radiative
  transfer of the cloudless atmosphere}.
\newblock \bibinfo{journal}{Renewable and Sustainable Energy Reviews}
  \bibinfo{volume}{188}, \bibinfo{pages}{113833}.
\newblock \URLprefix
  \url{https://linkinghub.elsevier.com/retrieve/pii/S1364032123006901},
  \DOIprefix\doi{10.1016/j.rser.2023.113833}.
\bibitem[{Ruiz-Arias and Gueymard(2018)}]{RUIZARIAS201810}
\bibinfo{author}{Ruiz-Arias, J.A.}, \bibinfo{author}{Gueymard, C.A.},
  \bibinfo{year}{2018}.
\newblock \bibinfo{title}{Worldwide inter-comparison of clear-sky solar
  radiation models: Consensus-based review of direct and global irradiance
  components simulated at the earth surface}.
\newblock \bibinfo{journal}{Solar Energy} \bibinfo{volume}{168},
  \bibinfo{pages}{10--29}.
\newblock \URLprefix
  \url{https://www.sciencedirect.com/science/article/pii/S0038092X18301257},
  \DOIprefix\doi{https://doi.org/10.1016/j.solener.2018.02.008}.
  \bibinfo{note}{advances in Solar Resource Assessment and Forecasting}.
\bibitem[{Sanfilippo et~al.(2016)Sanfilippo, Martin-Pomares, Mohandes,
  Perez-Astudillo and Bachour}]{sanfilippo2016adaptive}
\bibinfo{author}{Sanfilippo, A.}, \bibinfo{author}{Martin-Pomares, L.},
  \bibinfo{author}{Mohandes, N.}, \bibinfo{author}{Perez-Astudillo, D.},
  \bibinfo{author}{Bachour, D.}, \bibinfo{year}{2016}.
\newblock \bibinfo{title}{An adaptive multi-modeling approach to solar
  nowcasting}.
\newblock \bibinfo{journal}{Solar Energy} \bibinfo{volume}{125},
  \bibinfo{pages}{77--85}.
\newblock \DOIprefix\doi{10.1016/J.SOLENER.2015.11.041}.
\bibitem[{Santamouris et~al.(1990)Santamouris, Tselepidaki and
  Dris}]{perez2017predictive}
\bibinfo{author}{Santamouris, M.}, \bibinfo{author}{Tselepidaki, I.},
  \bibinfo{author}{Dris, N.}, \bibinfo{year}{1990}.
\newblock \bibinfo{title}{Evaluation of models to predict solar radiation on
  tilted surfaces for the mediterranean region}.
\newblock \bibinfo{journal}{Solar \& Wind Technology} \bibinfo{volume}{7},
  \bibinfo{pages}{585--589}.
\newblock \DOIprefix\doi{10.1016/0741-983X(90)90067-C}.
\bibitem[{Shan et~al.(2022)Shan, Li, Ding, Wang, Zhang and
  Wei}]{shan2022ensemble}
\bibinfo{author}{Shan, S.}, \bibinfo{author}{Li, C.}, \bibinfo{author}{Ding,
  Z.}, \bibinfo{author}{Wang, Y.}, \bibinfo{author}{Zhang, K.},
  \bibinfo{author}{Wei, H.}, \bibinfo{year}{2022}.
\newblock \bibinfo{title}{Ensemble learning based multi-modal intra-hour
  irradiance forecasting}.
\newblock \bibinfo{journal}{Energy Convers Manag} \bibinfo{volume}{270},
  \bibinfo{pages}{116206}.
\newblock \DOIprefix\doi{10.1016/J.ENCONMAN.2022.116206}.
\bibitem[{Silva et~al.(2021)Silva, Zarpelão, Cano and Junior}]{Silva2021}
\bibinfo{author}{Silva, R.P.}, \bibinfo{author}{Zarpelão, B.},
  \bibinfo{author}{Cano, A.}, \bibinfo{author}{Junior, S.B.},
  \bibinfo{year}{2021}.
\newblock \bibinfo{title}{Time series segmentation based on stationarity
  analysis to improve new samples prediction}.
\newblock \bibinfo{journal}{Sensors (Basel, Switzerland)} \bibinfo{volume}{21}.
\newblock \DOIprefix\doi{10.3390/s21217333}.
\bibitem[{Sobri et~al.(2018)Sobri, Koohi-Kamali and Rahim}]{sharma2020review}
\bibinfo{author}{Sobri, S.}, \bibinfo{author}{Koohi-Kamali, S.},
  \bibinfo{author}{Rahim, N.A.}, \bibinfo{year}{2018}.
\newblock \bibinfo{title}{Solar photovoltaic generation forecasting methods: A
  review}.
\newblock \bibinfo{journal}{Energy Conversion and Management}
  \bibinfo{volume}{156}, \bibinfo{pages}{459--497}.
\newblock \DOIprefix\doi{10.1016/j.enconman.2017.11.019}.
\bibitem[{Vicente-Serrano et~al.(2022)Vicente-Serrano, Domínguez-Castro, Reig,
  Beguería, Tomas-Burguera, Latorre, Peña-Angulo, Noguera, Rabanaque, Luna,
  Morata and {El Kenawy}}]{carrasco2017agroclimatic}
\bibinfo{author}{Vicente-Serrano, S.}, \bibinfo{author}{Domínguez-Castro, F.},
  \bibinfo{author}{Reig, F.}, \bibinfo{author}{Beguería, S.},
  \bibinfo{author}{Tomas-Burguera, M.}, \bibinfo{author}{Latorre, B.},
  \bibinfo{author}{Peña-Angulo, D.}, \bibinfo{author}{Noguera, I.},
  \bibinfo{author}{Rabanaque, I.}, \bibinfo{author}{Luna, Y.},
  \bibinfo{author}{Morata, A.}, \bibinfo{author}{{El Kenawy}, A.},
  \bibinfo{year}{2022}.
\newblock \bibinfo{title}{A near real-time drought monitoring system for spain
  using automatic weather station network}.
\newblock \bibinfo{journal}{Atmospheric Research} \bibinfo{volume}{271},
  \bibinfo{pages}{106095}.
\newblock \DOIprefix\doi{10.1016/j.atmosres.2022.106095}.
\bibitem[{Voyant et~al.(2022a)Voyant, Lauret, Notton, Duchaud, Garcia-Gutierrez
  and Faggianelli}]{10.1063/5.0128131}
\bibinfo{author}{Voyant, C.}, \bibinfo{author}{Lauret, P.},
  \bibinfo{author}{Notton, G.}, \bibinfo{author}{Duchaud, J.L.},
  \bibinfo{author}{Garcia-Gutierrez, L.}, \bibinfo{author}{Faggianelli, G.A.},
  \bibinfo{year}{2022}a.
\newblock \bibinfo{title}{Complex-valued time series based solar irradiance
  forecast}.
\newblock \bibinfo{journal}{Journal of Renewable and Sustainable Energy}
  \bibinfo{volume}{14}, \bibinfo{pages}{066502}.
\newblock \DOIprefix\doi{10.1063/5.0128131}.
\bibitem[{Voyant et~al.(2022b)Voyant, Notton, Duchaud,
  Garc{\'i}a~Guti{\'e}rrez, Bright and Yang}]{voyant2017benchmarks}
\bibinfo{author}{Voyant, C.}, \bibinfo{author}{Notton, G.},
  \bibinfo{author}{Duchaud, J.L.}, \bibinfo{author}{Garc{\'i}a~Guti{\'e}rrez,
  L.A.}, \bibinfo{author}{Bright, J.M.}, \bibinfo{author}{Yang, D.},
  \bibinfo{year}{2022}b.
\newblock \bibinfo{title}{Benchmarks for solar radiation time series
  forecasting}.
\newblock \bibinfo{journal}{Renewable Energy} \bibinfo{volume}{113},
  \bibinfo{pages}{102--114}.
\newblock \DOIprefix\doi{10.1016/j.renene.2022.04.065}.
\bibitem[{Voyant et~al.(2017)Voyant, Notton, Kalogirou, Nivet, Paoli, Motte and
  Fouilloy}]{voyant2017machine}
\bibinfo{author}{Voyant, C.}, \bibinfo{author}{Notton, G.},
  \bibinfo{author}{Kalogirou, S.}, \bibinfo{author}{Nivet, M.L.},
  \bibinfo{author}{Paoli, C.}, \bibinfo{author}{Motte, F.},
  \bibinfo{author}{Fouilloy, A.}, \bibinfo{year}{2017}.
\newblock \bibinfo{title}{Machine learning methods for solar radiation
  forecasting: A review}.
\newblock \bibinfo{journal}{Renewable Energy} \bibinfo{volume}{105},
  \bibinfo{pages}{569--582}.
\newblock \DOIprefix\doi{10.1016/j.renene.2016.12.095}.
\bibitem[{Wang et~al.(2022)Wang, Zhang, Lin, Wang and Zhu}]{Wang2022}
\bibinfo{author}{Wang, K.}, \bibinfo{author}{Zhang, Y.}, \bibinfo{author}{Lin,
  F.}, \bibinfo{author}{Wang, J.}, \bibinfo{author}{Zhu, M.},
  \bibinfo{year}{2022}.
\newblock \bibinfo{title}{Nonparametric probabilistic forecasting for wind
  power generation using quadratic spline quantile function and autoregressive
  recurrent neural network}.
\newblock \bibinfo{journal}{IEEE Transactions on Sustainable Energy}
  \bibinfo{volume}{13}, \bibinfo{pages}{1930--1943}.
\newblock \DOIprefix\doi{10.1109/TSTE.2022.3175916}.
\bibitem[{Yang(2019)}]{YANG2019981}
\bibinfo{author}{Yang, D.}, \bibinfo{year}{2019}.
\newblock \bibinfo{title}{Making reference solar forecasts with climatology,
  persistence, and their optimal convex combination}.
\newblock \bibinfo{journal}{Solar Energy} \bibinfo{volume}{193},
  \bibinfo{pages}{981--985}.
\newblock \DOIprefix\doi{10.1016/j.solener.2019.10.006}.
\bibitem[{Yang et~al.(2022)Yang, Wang, Gueymard, Hong, Kleissl, Huang, Perez,
  Perez, Bright, Xia, {van der Meer} and Peters}]{yang2018review}
\bibinfo{author}{Yang, D.}, \bibinfo{author}{Wang, W.},
  \bibinfo{author}{Gueymard, C.A.}, \bibinfo{author}{Hong, T.},
  \bibinfo{author}{Kleissl, J.}, \bibinfo{author}{Huang, J.},
  \bibinfo{author}{Perez, M.J.}, \bibinfo{author}{Perez, R.},
  \bibinfo{author}{Bright, J.M.}, \bibinfo{author}{Xia, X.},
  \bibinfo{author}{{van der Meer}, D.}, \bibinfo{author}{Peters, I.M.},
  \bibinfo{year}{2022}.
\newblock \bibinfo{title}{A review of solar forecasting, its dependence on
  atmospheric sciences and implications for grid integration: Towards carbon
  neutrality}.
\newblock \bibinfo{journal}{Renewable and Sustainable Energy Reviews}
  \bibinfo{volume}{161}, \bibinfo{pages}{112348}.
\newblock \DOIprefix\doi{10.1016/j.rser.2022.112348}.
\bibitem[{Zamo and Naveau(2018)}]{Zamo2018}
\bibinfo{author}{Zamo, M.}, \bibinfo{author}{Naveau, P.}, \bibinfo{year}{2018}.
\newblock \bibinfo{title}{Estimation of the continuous ranked probability score
  with limited information and applications to ensemble weather forecasts}.
\newblock \bibinfo{journal}{Mathematical Geosciences} \bibinfo{volume}{50},
  \bibinfo{pages}{209--234}.
\newblock \DOIprefix\doi{10.1007/s11004-017-9709-7}.
\bibitem[{Zheng et~al.(2023)Zheng, Ge, Muhsen, Wang, Elkamchouchi, Ali and
  Ali}]{lukasik2020ridge}
\bibinfo{author}{Zheng, Y.}, \bibinfo{author}{Ge, Y.}, \bibinfo{author}{Muhsen,
  S.}, \bibinfo{author}{Wang, S.}, \bibinfo{author}{Elkamchouchi, D.H.},
  \bibinfo{author}{Ali, E.}, \bibinfo{author}{Ali, H.E.}, \bibinfo{year}{2023}.
\newblock \bibinfo{title}{New ridge regression, artificial neural networks and
  support vector machine for wind speed prediction}.
\newblock \bibinfo{journal}{Advances in Engineering Software}
  \bibinfo{volume}{179}, \bibinfo{pages}{103426}.
\newblock \DOIprefix\doi{10.1016/j.advengsoft.2023.103426}.

\end{thebibliography}
\end{document}